%% file: main.tex
\documentclass{article}

\usepackage[T1]{fontenc}  %
\usepackage{lmodern}      %
\usepackage{amsmath}
\usepackage{amssymb}
\usepackage{graphicx}
\usepackage{array}
\usepackage{booktabs}
\usepackage[section]{placeins}  %
\usepackage[font=small,labelfont=bf]{caption}
\usepackage[round,semicolon]{natbib}  %
\usepackage{xcolor}
\usepackage{hyperref}
\usepackage{doi}  %
\definecolor{linkblue}{rgb}{0 0.4235 0.6392}
\hypersetup{colorlinks, linkcolor=linkblue, citecolor=linkblue, urlcolor=linkblue,
  pdftitle={Translating KV Caches Between Hybrid-Attention Language Models},
  pdfauthor={Povilas Kvedaras, Tomas Rasymas}}
\usepackage{tabularx}     %
\usepackage{adjustbox}    %
\usepackage{pifont}       %


\newcolumntype{L}[1]{>{\raggedright\arraybackslash}p{#1}}
\definecolor{brandprimary}{HTML}{673DE6}  %
\newcommand{\cpone}{\ding{172}}
\newcommand{\cptwo}{\ding{173}}
\newcommand{\cpthree}{\ding{174}}
\newcommand{\cpr}{\textcolor{brandprimary}{\ding{202}}}
\newcommand{\pairA}{Pair~A}
\newcommand{\pairB}{Pair~B}
\newcommand{\pairC}{Pair~C}
\newcommand{\fwd}{$\rightarrow$}
\newcommand{\bidir}{$\leftrightarrow$}
\newcommand{\pmse}[1]{{\scriptsize$\pm$#1}}
\newcommand{\ci}[2]{{\scriptsize[#1,\,#2]}}

\title{Translating KV Caches Between\\ Hybrid-Attention Language Models:\\ MLA Latents, Linear-Attention States\\
and Sparse-Attention Indexers}
\author{Povilas Kvedaras\thanks{Corresponding author:
  \href{mailto:povilas.kvedaras@hostinger.com}{povilas.kvedaras@hostinger.com}.}\qquad Tomas Rasymas\\[2pt]
  Hostinger Research}
\date{}

\begin{document}
\maketitle

\begin{abstract}
A model that takes over another model's conversation must normally re-read the whole history; a text hand-off
shortens the re-read only by dropping content.
KV-cache translation instead maps the source's cache into the target's own format, so the target continues from the
full context without re-reading it.
It works for grouped-query attention (GQA), but open-weights frontier models now cache compressed latents
(MLA), sparse-attention indexer keys and fixed-size linear-attention states.
We replicate KV-Lingo on a GQA pair and then give, for each of these objects, where to capture and inject it and a
mechanism that reaches native quality.
We study two hybrid pairs at 16--48B scale in both directions, Kimi-VL-A3B $\leftrightarrow$ Kimi-Linear-48B
(MLA $\leftrightarrow$ MLA + Kimi Delta Attention) and Qwen3.5-27B $\leftrightarrow$ Qwen3.8-27B (Gated DeltaNet
hybrids), plus a sparse-attention target (Qwen3.8-27B $\rightarrow$ Qwen3.8-Flash-Next).
MLA latents translate with token-wise linear maps; recurrent states are copied with a closed-form correction or
synthesised by replaying the target's own recurrence.
Exact long-range recall needs token identity: decoding token ids from the source's layer-0 MLA latent raises needle
recall from 0.09 to 0.99 at 2--32K and to 0.83--1.00 at 64K, eight times the longest training prompt.
The hybrid $\rightarrow$ full-MLA direction is within the paired confidence interval of native quality on every task
metric, and same-shape GDN hybrids reach the native level in both directions (apart from 6\% refusals inherited from
the source), with time to first token 18.7 and 46.8 times shorter than a re-read at 64K tokens.
Segment-wise multi-turn transfer stays at the re-prefill level; naive re-translation collapses.
On the sparse-attention target, predicted indexer keys recall the needle as well as the target's own keys (one seed).
We state as a hypothesis how the recipes transfer to 300B+ hybrid flagships.
\end{abstract}

\begin{quote}\small
\end{quote}

\input{intro}
\input{background}
\input{method}
\input{setup}
\input{replication}
\input{results_quality}
\input{results_speed}
\input{results_multiturn}
\input{analysis}
\input{frontier}
\input{limitations}

\FloatBarrier

\section*{Data and code availability}
The translator code and the trained translator weights are kept in Hostinger's internal research repositories and are
not publicly released. The source and target models are openly released checkpoints and are subject to their
licences; the evaluation data are public benchmarks (CoQA, ARC, GSM8K) and synthetic needle-in-a-haystack prompts.

\section*{Competing interests}
The authors are employees of Hostinger, which uses language models in its products. The authors declare no other
competing interests.

\section*{Use of generative AI}
An LLM agent helped to implement the translator candidates code under a sealed evaluator. The authors designed the
study, checked every reported result against its result file, reviewed and edited all co-generated text, and took full
responsibility for the content of the paper.

\bibliographystyle{plainnat}
\bibliography{references}

\input{appendix.tex}

\end{document}

%% file: intro.tex
\section{Introduction}
\label{sec:intro}

When a conversation moves from one language model to another, for example from a small model to a larger one or
between two generations of a model family, the second model normally re-reads (prefills) the whole history.
KV-cache translation avoids this.
A small learned map converts the source model's key--value cache into the target model's cache format, and the target
continues from the translated cache as if it had read the prompt itself.
KV-Lingo~\citep{castin2026kvlingo} showed that, for two grouped-query attention (GQA) models with the same tokenizer,
per-layer linear maps fitted in closed form and refined by self-distillation give the target its native quality at a
fraction of the prefill cost, and that translating each conversation segment once keeps multi-turn quality at the
re-prefill level.

\begin{table}[t]
\centering\scriptsize
\setlength{\tabcolsep}{2.6pt}
\caption{Cache objects of recent open-weights releases, newest first within each family (configurations read from
each model's \texttt{config.json} on Hugging Face, 3 Nov 2026; $\star$ = the family's most downloaded repository in
the last month). K/V = per-token keys and values; MLA = compressed latent per token; Idx = sparse-attention indexer
key per token or block; State = linear-attention or state-space recurrent state (bf16 size per prompt where a pair
rehearses it); Other = sliding window or block-compressed shared K/V. Last column: the pair of this paper with the
same objects (close = same object types, shapes, tokenizer setting and layer ratio; partial = an object or a ratio
differs; Figure~\ref{fig:frontier} compares cache bytes per token).}
\label{tab:evolution}
\resizebox{\textwidth}{!}{%
\begin{tabularx}{5.5in}{@{}p{2.55cm}p{1.15cm}p{1.35cm}ccccc>{\raggedright\arraybackslash}Xp{1.75cm}@{}}
\toprule
Family / model & Released & B total / act. & K/V & MLA & Idx & State & Other & Layers & Our pair (match) \\
\midrule
DeepSeek-V4-Flash$^\star$ / Pro~\citep{deepseek2026v4} & Apr 2026 & 284/13; 1600/49 & & ($\checkmark$) & \checkmark & & \checkmark & one 512-d KV head, block-compressed (CSA/HCA), DSA indexer, 128-token window & not rehearsed \\
DeepSeek-V3.2~\citep{deepseek2025v32} & Dec 2025 & 685 / 37 & & \checkmark & \checkmark & & & 61 MLA + DSA (top-2048) & origin of DSA \\
\addlinespace[2pt]
GLM-5.3-Flash$^\star$ & Aug 2026 & 320 / 18 & & \checkmark & \checkmark & \checkmark & & 34 KDA + 11 DSA-MLA, pooled indexer; 142\,MiB & A target (partial: DSA) \\
GLM-5.2 / 5.3 & Jun--Aug 2026 & 744 / 40 & & \checkmark & \checkmark & & & 78 MLA + DSA, 21 shared indexers~\citep{bai2026indexcache} & A source (partial: DSA) \\
GLM-5~\citep{glm2026glm5} & Feb 2026 & 744 / 40 & & \checkmark & \checkmark & & & 78 MLA + DSA & \\
GLM-4.7 & Dec 2025 & 355 / 32 & \checkmark & & & & & 92 GQA, 8 kv heads & \\
\addlinespace[2pt]
Kimi-K3$^\star$~\citep{moonshot2026kimik3} & Jul 2026 & 2800 / 104 & & \checkmark & & \checkmark & & 69 KDA (96 heads) + 24 gated MLA (NoPE); 433\,MiB & A target (close; 10.3$\times$ state) \\
Kimi-K2.6 / K2-Thinking & Nov 2025--Apr 2026 & 1000 / 32 & & \checkmark & & & & 61 MLA & A source (close) \\
Kimi-Linear-48B~\citep{kimi2025linear} & Oct 2025 & 48 / 3 & & \checkmark & & \checkmark & & 20 KDA + 7 MLA; 41.9\,MiB & A target (itself) \\
\addlinespace[2pt]
Qwen3.8-Flash-Next$^\star$~\citep{qiu2026qwen38next} & Aug 2026 & 125 / 6 & \checkmark & & \checkmark & \checkmark & & 36 GDN + 12 QSA (gated attention + block indexer); 111\,MiB & B; C target (itself) \\
Qwen3.8-2.4T-A95B & Aug 2026 & 2400 / 95 & \checkmark & & & \checkmark & & 69 GDN (128 heads) + 23 GQA; 563\,MiB & B (partial: 128 vs 48 heads) \\
Qwen3.5-397B~\citep{qwen2026qwen35} & Feb 2026 & 397 / 17 & \checkmark & & & \checkmark & & 45 GDN + 15 gated attention & B \\
Qwen3-Next-80B & Sep 2025 & 80 / 3 & \checkmark & & & \checkmark & & 36 GDN + 12 gated attention & B \\
Qwen3~\citep{qwen2025qwen3} & May 2025 & 0.6--235 & \checkmark & & & & & GQA & Part 1 (itself) \\
\addlinespace[2pt]
Mistral Small 4$^\star$ / Large 3 & Dec 2025--Mar 2026 & 119/6.5; 675/41 & & \checkmark & & & & 36 / 61 MLA & A source (MLA only) \\
\addlinespace[2pt]
Nemotron 3 Nano$^\star$ / Super / Ultra~\citep{nvidia2025nemotron3} & Dec 2025--Jun 2026 & 30/3.5; 120/12; 550/55 & \checkmark & & & \checkmark & & Mamba-2 with 6--12 GQA layers (2 kv heads) & not rehearsed (Mamba-2) \\
\addlinespace[2pt]
MiniMax-M3~\citep{lai2026msa} & Jun 2026 & 428 / 23 & \checkmark & & \checkmark & & & 60 GQA + block-sparse indexer (MSA) & Part 1 + C (partial) \\
MiniMax-M2.5$^\star$ & Feb 2026 & 229 / 10 & \checkmark & & & & & 62 GQA & Part 1 \\
\addlinespace[2pt]
Meta Muse Glimmer$^\star$ & Aug 2026 & 30 / 30 & \checkmark & & & & \checkmark & 39 sliding-window (2,048) + 13 global NoPE GQA layers, 2 kv heads & Part 1 (partial: window) \\
\bottomrule
\end{tabularx}}
\end{table}

\paragraph{Why translate the cache.}
Model switches are routine: routers send each turn to a cheaper or a stronger model~\citep{ong2024routellm}, agent
pipelines hand one conversation between models, and serving fleets move users to a new model generation. Today the
receiving model either re-reads the full history, whose cost grows with the prompt length and the model's size, or it
receives a text summary of it, which is shorter to read but drops content the summary did not keep and still has to
be prefilled. Latent agent-to-agent communication~\citep{fu2026c2c,rossi2026lcf} passes richer information but keeps
both models in the loop and does not replace the target's prefill. A translated cache gives the target the complete
history in its own format, at a cost that grows with the prompt length but not with the target's size, and with no
text in between: in the two directions we timed at 64K tokens the target starts answering 18.7 and 46.8 times
sooner than after a re-read, with task quality within the paired confidence interval of its native level.

\paragraph{Frontier caches are no longer plain GQA.}
Table~\ref{tab:evolution} lists the cache objects of recent releases of the seven open-weights families with the
most-downloaded models on Hugging Face. Within one year every family except Meta's has moved its newest model away
from a plain per-token K/V cache, and the newest models of DeepSeek, GLM, Kimi and Qwen combine two or three different
objects in one stack. Multi-head latent attention (MLA)~\citep{deepseek2024v2} caches one 512-d latent plus a 64-d
rotary key per token and layer instead of per-head keys and values. Sparse-attention indexers such as DeepSeek Sparse
Attention (DSA)~\citep{deepseek2025v32} add a small \emph{indexer key} per token, and each query attends only to the
top-$k$ tokens the indexer selects. Linear-attention or state-space layers such as Kimi Delta Attention
(KDA)~\citep{kimi2025linear}, Gated DeltaNet (GDN)~\citep{yang2025gdn} and Mamba-2~\citep{dao2024mamba2} keep no
per-token cache at all, only a fixed-size recurrent state. Each of these objects breaks an assumption of the GQA
recipe: a token-wise map cannot produce a recurrent state, an MLA latent is normalised and partly rotated before it
is cached, and an indexer key decides which tokens attention reads at all. Which translation mechanism works for each
object, and where in the layer it should read and write (its \emph{capture point}, \S\ref{sec:background},
Figures~\ref{fig:legend}--\ref{fig:lin}), is open.

\begin{figure}[t]
\centering
\includegraphics[width=\textwidth]{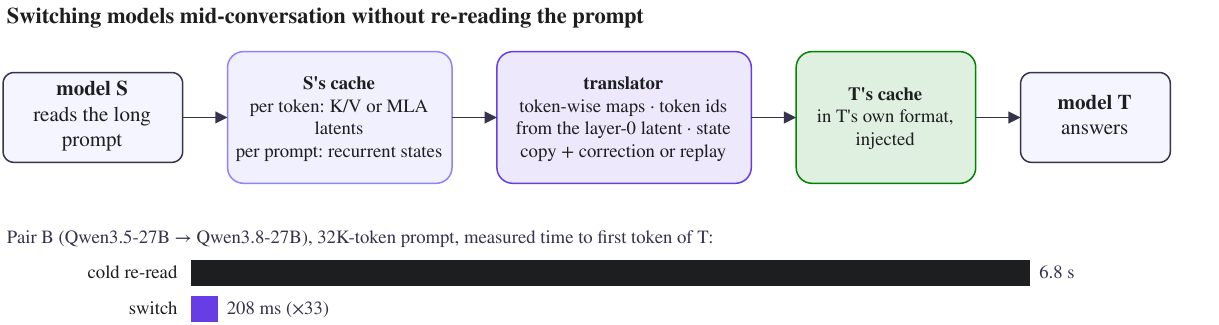}
\caption{Switching models mid-conversation without re-reading the prompt. Model S has read the prompt; a translator
maps S's cache (per-token K/V or MLA latents and per-prompt recurrent states) into T's format, and T answers. Inset:
\pairB{} at 32K tokens, time to T's first token, cold re-read vs switch (6,823 vs 208\,ms, $\times$32.8).}
\label{fig:teaser}
\end{figure}

\paragraph{This paper.}
We first replicate KV-Lingo on a GQA pair (Qwen3-8B \bidir{} Qwen3-4B). We then ask, for each cache object, which
mechanism reproduces the GQA-level result on hybrid models, using pairs that contain the flagships' cache objects at
16--125B scale (last column of Table~\ref{tab:evolution}; \S\ref{sec:setup} discusses where they differ):
\begin{itemize}
\item \textbf{\pairA}: Kimi-VL-A3B (27 MLA layers) \bidir{} Kimi-Linear-48B-A3B (20 KDA + 7 MLA layers).
\item \textbf{\pairB}: Qwen3.5-27B \bidir{} Qwen3.8-27B, two GDN + gated-attention hybrids with identical cache
  geometry.
\item \textbf{\pairC}: Qwen3.8-27B \fwd{} Qwen3.8-Flash-Next (36 GDN + 12 QSA layers), a target with a
  sparse-attention indexer, run at smaller scale (one direction, one seed).
\end{itemize}
We report quality, switch speed and multi-turn behaviour in both directions for Pairs A and B, an explicit account of
where each object is captured and injected, and the indexer rehearsal on \pairC{}. Figure~\ref{fig:teaser} shows the
setting.

\paragraph{What is new.}
Prior cache translators, including KV-Lingo and the cross-model methods of \S\ref{sec:related}, map the per-token
K/V cache of standard attention; one concurrent preprint~\citep{villani2026latentport} also transfers a GDN state on one pair in one direction. We
treat every cache object of current hybrid models (MLA latents, KDA/GDN states and sparse-attention indexer keys) and
show that each needs its own mechanism and its own capture point: token-wise maps where the object is per token, a
copy or a replay of the target's own recurrence where it is a state, and token identity, read back from the source's
own MLA cache, where exact recall depends on it.

\paragraph{Contributions.}
\begin{enumerate}
\item \textbf{A capture/injection map for every cache object} (\S\ref{sec:capture}): GQA K/V, MLA latent and rotary
  key, KDA/GDN state and conv window, and sparse-attention indexer key. We give the engine-native object, the point we
  capture, where we write, and an ablation of capture point \cpone{} (pre-norm) vs \cpthree{} (the engine's own cache)
  (\S\ref{sec:cp3}).
\item \textbf{Per-object recipes with measured evidence} (Table~\ref{tab:recipes}): token-wise maps for MLA latents;
  token identity, decoded from the source's own layer-0 MLA latent, for exact long-range recall; state copy with a
  closed-form bilinear correction when state geometries match; replay of the target's own recurrence when the source
  has no state.
\item \textbf{Both directions on both hybrid pairs}, with three seeds for the headline forward rows. The
  hybrid\,\fwd\,full-MLA direction (a Flash hybrid handing over to its full-MLA flagship) is the easier one.
\item \textbf{Speed and multi-turn} on the same GPUs and kernels as the cold prefill, and a two-direction
  reproduction of KV-Lingo's naive-re-translation collapse on a hybrid pair.
\item \textbf{A rehearsal of sparse-attention indexer transfer} (\pairC{}, one seed, 400 and 1000 steps) and a
  labelled argument for how the recipes transfer to 300B+ flagships (\S\ref{sec:frontier}).
\end{enumerate}

\begin{table}[t]
\centering\small
\caption{Recipes per cache object, with the evidence each rests on. \cpone{} and \cpthree{} are capture points
(raw projection output and the engine's cached tensor; Figures~\ref{fig:legend}--\ref{fig:lin}). ``Native level'' = the paired 95\% CI
against the target reading its own cache includes 0, or the gap is within the SE.}
\label{tab:recipes}
\begin{tabularx}{\linewidth}{@{}p{2.2cm}p{4.1cm}X@{}}
\toprule
Object & Recipe & Evidence (section) \\
\midrule
GQA K/V & token-wise linear map per layer, closed form + distillation; capture \cpone{} or \cpthree{} & Part 1 at native
judge level, needle 1.00 at 2--8K (\S\ref{sec:replication}); \pairB{} \cpthree{} ties \cpone{} to 128K (\S\ref{sec:cp3}) \\
MLA latent + rotary key & token-wise map on the pre-norm, un-rotated object \cpone{}; inject through the target's own
norm (and RoPE, if any) & \pairA{} reverse within the CI of native on every task metric (one seed), needle 1.00 to
64K (\S\ref{sec:quality}) \\
Token identity & decode token ids from the source's layer-0 MLA latent (or pass ids) and feed the target's embedding
to the maps & needle 0.09\,\fwd\,0.99 (\pairA{} forward), 64K 0.00\,\fwd\,0.83--1.00 (\S\ref{sec:tokid}) \\
Recurrent state, same geometry & copy + closed-form bilinear correction $A S B$ per head; O(1) in length & \pairB{} at
native level both ways, needle 1.00 at 128K, $\times$46.8 at 64K (\S\ref{sec:quality}, \S\ref{sec:speed}) \\
Recurrent state, no source state & replay the target's own recurrence over predicted per-token inputs; O($n$) &
\pairA{} forward; zero state fails (val KL 0.082\,\fwd\,0.344) (\S\ref{sec:analysis}) \\
Sparse indexer key & predict from translated K + token embedding, or attend densely over the translated prefix &
\pairC{}: pooled needle 0.87/0.84 with predicted keys vs 0.88/0.85 with native keys (target 0.91);
CoQA F1 0.80/0.77 vs 0.77; 400/1000 steps, one seed (\S\ref{sec:dsa}) \\
Multi-turn & segment-wise transfer (translate each segment once); optional native refresh & both pairs at the
re-prefill level on content; naive collapses on \pairA{} (\S\ref{sec:multiturn}) \\
\bottomrule
\end{tabularx}
\end{table}

%% file: background.tex
\section{Background: cache formats as objects}
\label{sec:background}

\begin{figure}[t]
\centering
\includegraphics[width=\textwidth]{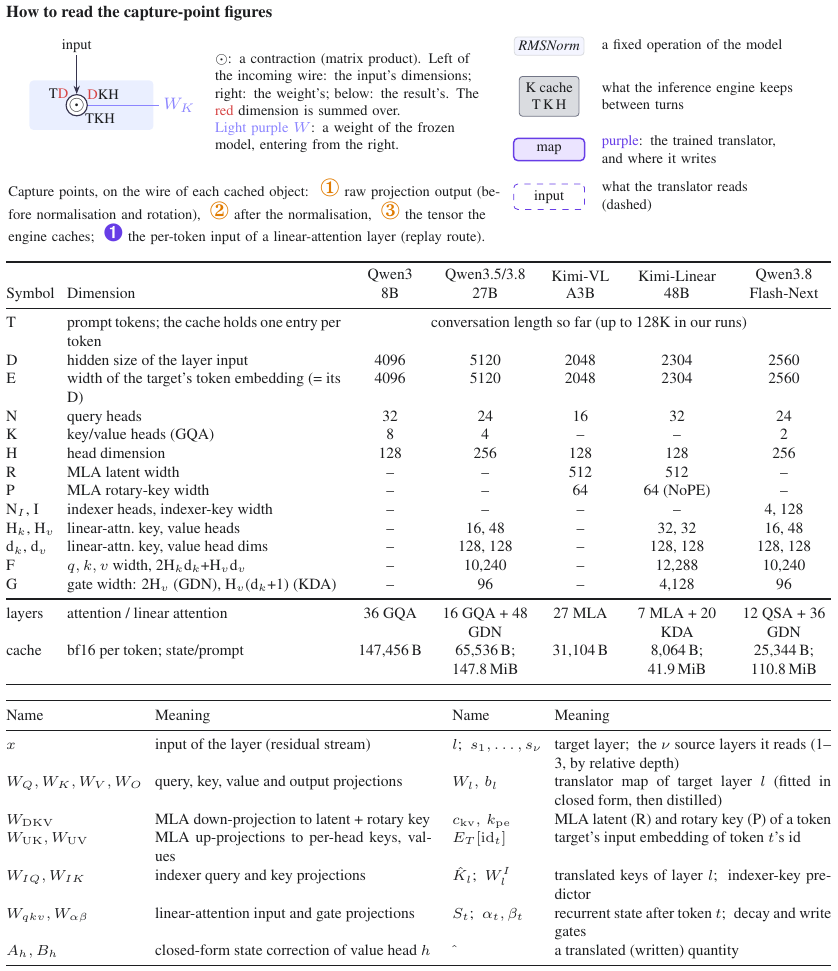}
\caption{How to read Figures~\ref{fig:gqa}--\ref{fig:lin}, which draw one attention layer of each variant (attention block
only) and where the translator reads and writes. Top: visual conventions and capture points. Middle: every dimension
letter with its value for our five models (cache sizes in bf16, MTP layers excluded; state sizes as measured). Bottom:
every named quantity.}
\label{fig:legend}
\end{figure}

\begin{figure}[t]
\centering
\includegraphics[width=\textwidth]{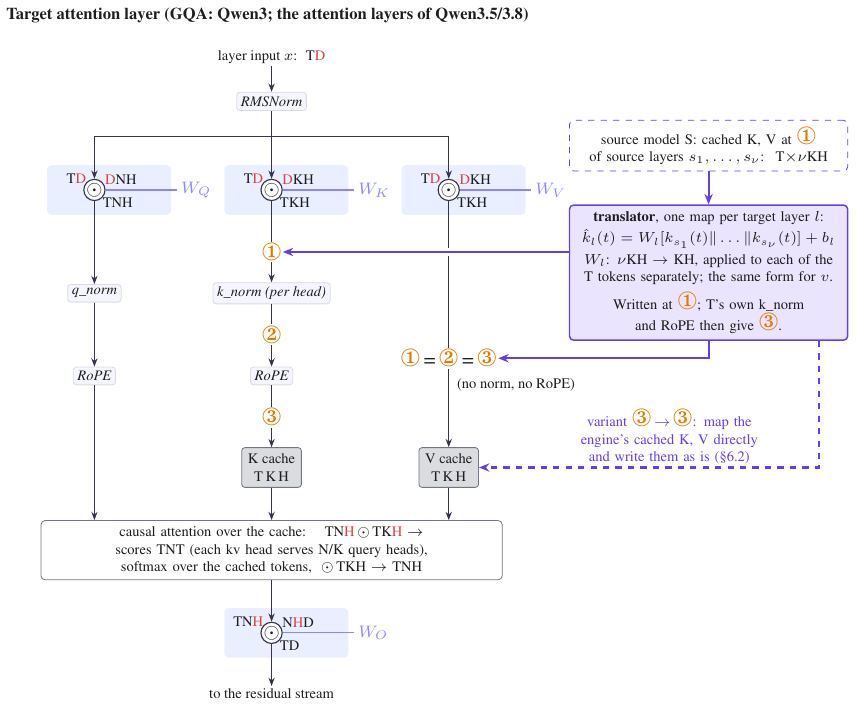}
\caption{GQA layer (Qwen3; the attention layers of Qwen3.5/3.8). The translator maps the source's keys and values at
\cpone{} of $\nu$ source layers, token by token, and writes them at \cpone{}; the target's own \texttt{k\_norm} and
RoPE then produce the cached keys \cpthree{}. Values are the same at all three points. Dashed: the \cpthree{}\,$\to$\,\cpthree{}
variant of \S\ref{sec:cp3}.}
\label{fig:gqa}
\end{figure}

\begin{figure}[t]
\centering
\includegraphics[width=\textwidth]{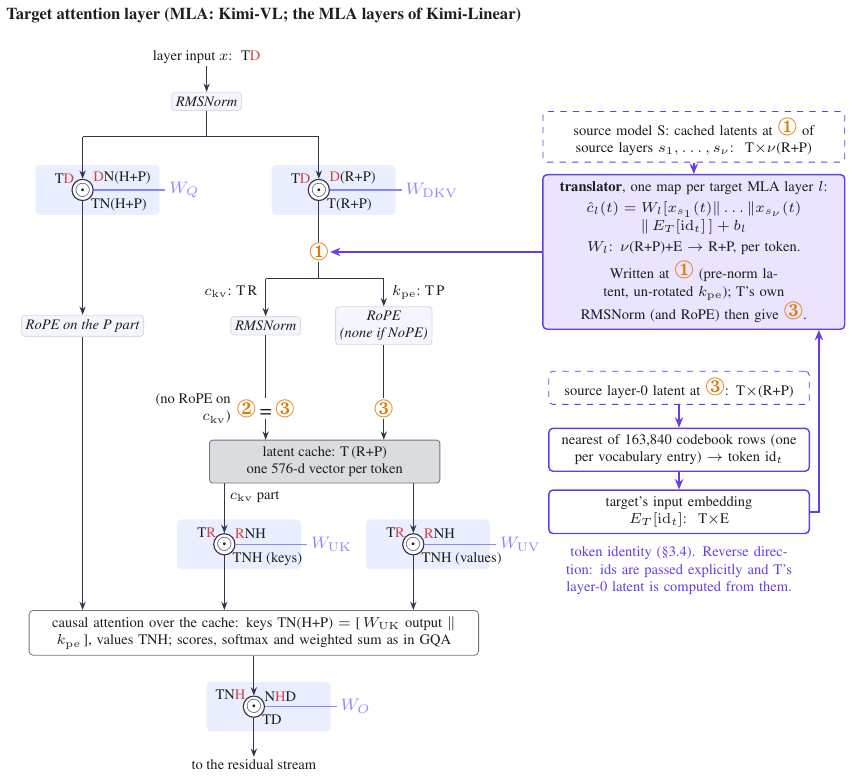}
\caption{MLA layer (Kimi-VL; the MLA layers of Kimi-Linear). The engine caches one 576-d vector per token (latent
$c_\mathrm{kv}$ and rotary key $k_\mathrm{pe}$) and up-projects it at attention time. The translator writes the pre-norm
latent and un-rotated rotary key at \cpone{}; its input is the source latents plus the target's embedding of each token
id, decoded from the source's layer-0 latent (\S\ref{sec:tokid-method}).}
\label{fig:mla}
\end{figure}

\begin{figure}[t]
\centering
\includegraphics[width=\textwidth]{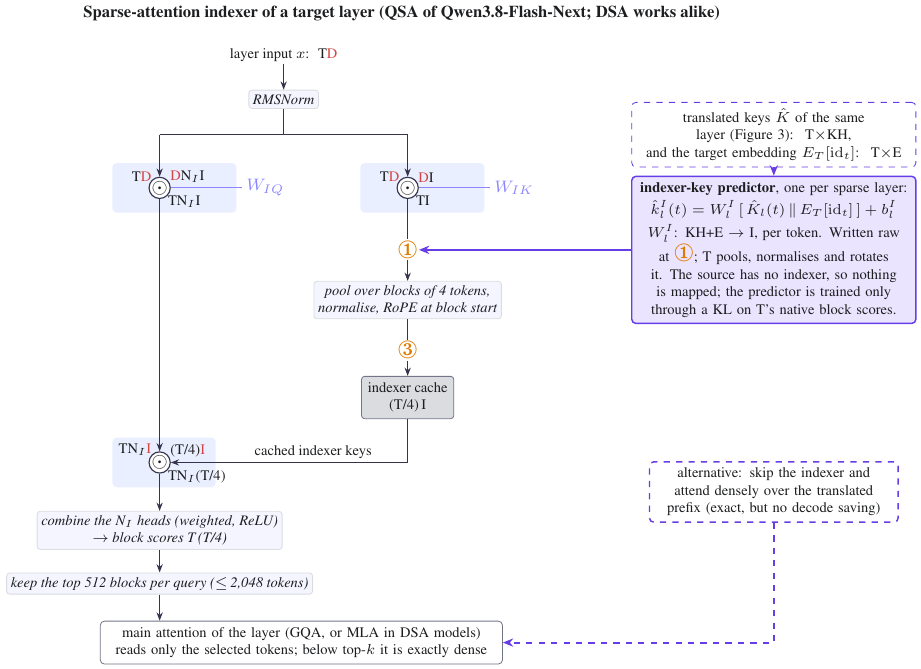}
\caption{Sparse-attention indexer (QSA of Qwen3.8-Flash-Next; DSA works alike). The indexer key decides which blocks
the main attention reads. Our source has no indexer, so a predictor writes the raw key at \cpone{} from the translated
keys and the token embedding, and the target pools, normalises and rotates it; the dashed alternative skips the
indexer and attends densely (\S\ref{sec:dsa}).}
\label{fig:idx}
\end{figure}

\begin{figure}[t]
\centering
\includegraphics[width=\textwidth]{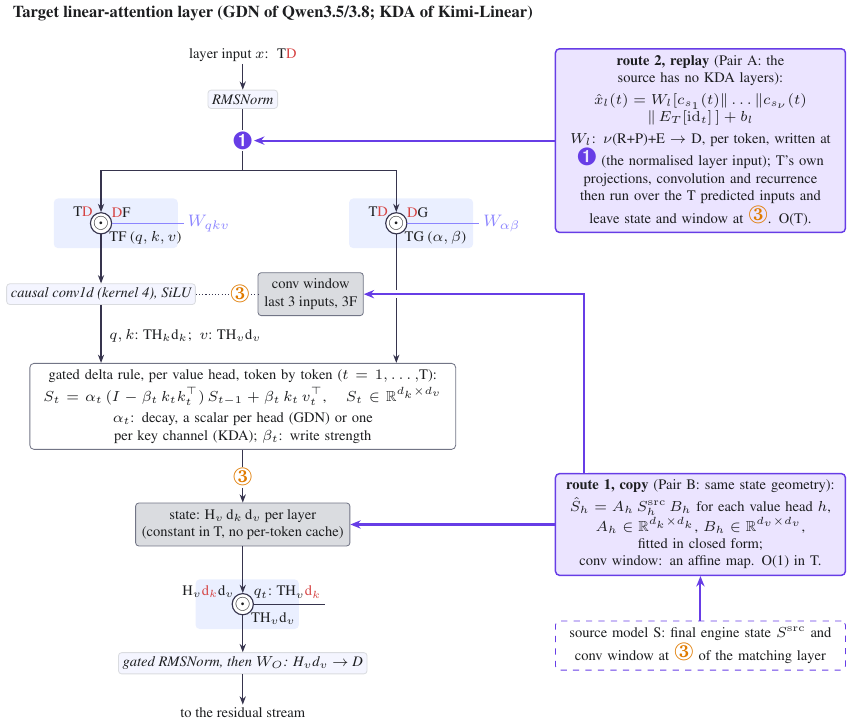}
\caption{Linear-attention layer (GDN of Qwen3.5/3.8; KDA of Kimi-Linear). The layer caches only a fixed-size state and a
conv window (\cpthree{}). Route~1 (\pairB{}) copies the source's state with a closed-form per-head correction, O(1) in
the prompt length; route~2 (\pairA{}, no source state) writes predicted per-token inputs at \cpr{} and replays the
target's own projections, convolution and recurrence over them, O($n$).}
\label{fig:lin}
\end{figure}

\paragraph{What a cache is.}
While a transformer reads a prompt, every attention layer stores something about each token so that later tokens can
attend to it; this store is the KV cache, and it is what a serving engine keeps between turns. Different attention
variants store different things. We call each kind of stored tensor a cache \emph{object} and describe it by its
\emph{granularity} (one entry per token, or one per prompt), its \emph{size}, and the \emph{operations} a translator
can apply to it (map it token-wise, copy it, replay it through the target's own layer, or compute it exactly from the
token ids). Figures~\ref{fig:gqa}--\ref{fig:lin} draw each object inside its attention layer, and
Figure~\ref{fig:legend} gives the symbols and the sizes for our models. GQA stores a key and a value per token and layer, so its per-token footprint is the largest (147\,KB for
Qwen3-8B) and the natural operation is a per-token linear map. MLA stores one compressed 576-d vector per token and
layer, five times less, and is also mapped token-wise; its first layer additionally identifies the token. DSA adds a
small indexer key per token that a translator must predict or map, or else fall back to dense attention.
Linear-attention layers (KDA, GDN) store nothing per token: one fixed-size matrix per layer summarises the whole
prompt, so the only options are to copy it (if the target has the same shape) or to rebuild it by running the
target's own recurrence. The rest of the paper gives one recipe per object.

\paragraph{How we measure fidelity.}
Throughout the paper the primary measure of a translated cache is a Kullback--Leibler (KL) divergence. Let $p_T(\cdot \mid
\mathcal{C}_T)$ be the target's next-token distribution when it reads its own cache of a prefix, and $p_T(\cdot \mid
\hat{\mathcal{C}}_T)$ the same model's distribution when it reads the translated cache instead. The \emph{validation
KL} is $\mathrm{KL}(p_T(\cdot \mid \mathcal{C}_T)\,\|\,p_T(\cdot \mid \hat{\mathcal{C}}_T))$ averaged over the
continuation tokens of held-out chat conversations (\S\ref{sec:stage2} gives the full expression); 0 means the target
cannot tell the translated cache from its own, and the same quantity is the training loss of Stage 2. We also report
the \emph{first-token KL}, the same divergence at the first token the target decodes. KL is a model-internal measure,
so we always pair it with task probes (CoQA, ARC, a needle recall test and a chat judge, \S\ref{sec:setup}); the
rehearsal in \S\ref{sec:dsa} shows a case where KL alone would have hidden a failure.

\paragraph{Capture points: where an object is read and written.}
A cached object passes through several transformations before the engine stores it, and a translator can read or write
it at any of them. Following KV-Lingo we name three points (Figure~\ref{fig:legend}): \cpone{} the raw output of the
projection that produces the object (before any normalisation or positional rotation), \cptwo{} after the
normalisation, and \cpthree{} the tensor the engine actually keeps. For a GQA key the three points are the
\texttt{k\_proj} output, the output of the per-head \texttt{k\_norm}, and the key after RoPE; values have no norm or
rotation, so their three points coincide (Figure~\ref{fig:gqa}). For an MLA layer the object is the compressed latent: \cpone{} is
the \texttt{kv\_a\_proj} output and \cpthree{} the normalised latent with its rotated key (Figure~\ref{fig:mla}). A sparse-attention
indexer key is captured raw (\cpone{}); the engine's pooled and rotated key (\cpthree{}) is a fixed function of it
(Figure~\ref{fig:idx}). For a linear-attention layer the analogue is the choice between copying the engine's state (\cpthree{}) and
rebuilding it from the layer's per-token inputs (Figure~\ref{fig:lin}). The point matters for two reasons: \cpone{} objects are position-free and on a common scale, so a
linear map fits them better, while \cpthree{} objects are what an unmodified engine exposes, so a translator at
\cpthree{} needs no model changes. \S\ref{sec:capture} gives the capture and injection point of every object we
translate, and \S\ref{sec:cp3} measures \cpone{} against \cpthree{}.

\paragraph{GQA (Figure~\ref{fig:gqa}).}
Each attention layer caches a key and a value per token and kv head. Keys are normalised (in Qwen3, a per-head
\texttt{k\_norm}) and rotated (RoPE) before caching; an inference engine such as vLLM~\citep{kwon2023vllm} stores the
\cpthree{} key. Values are the \texttt{v\_proj} output at every point.

\paragraph{MLA (Figure~\ref{fig:mla}).}
Multi-head latent attention~\citep{deepseek2024v2} projects the hidden state to a compressed latent $c_\mathrm{kv}$
(512) and a decoupled rotary key $k_\mathrm{pe}$ (64). The engine caches $\mathrm{RMSNorm}(c_\mathrm{kv})\,\|\,
\mathrm{RoPE}(k_\mathrm{pe})$ and up-projects the latent to per-head keys and values at attention time. Kimi-Linear's MLA
layers use no positional encoding (NoPE): $k_\mathrm{pe}$ is cached un-rotated. GLM-5.3-Flash has no rotary part at all.

\paragraph{Sparse attention indexers (Figure~\ref{fig:idx}).}
DSA~\citep{deepseek2025v32} keeps the MLA cache and adds a lightning indexer: per token a small key (128-d), stored in
low precision. At decode, each query scores the cached indexer keys and the main attention reads only the top-$k$
tokens (2,048 for GLM-5.x and DeepSeek-V3.2). GLM-5.2/5.3 share each indexer across 4 layers (21 of 78 layers store a
key~\citep{bai2026indexcache}); GLM-5.3-Flash and Qwen3.8-Flash-Next (QSA) pool indexer keys over blocks of 4 tokens.
Below top-$k$, sparse attention is exactly dense.

\paragraph{Linear attention (KDA, GDN; Figure~\ref{fig:lin}).}
Gated DeltaNet~\citep{yang2025gdn} and Kimi Delta Attention~\citep{kimi2025linear} replace the per-token cache by a
per-head matrix state updated by a gated delta rule, plus a short causal convolution whose last inputs are cached
(conv window). The state has constant size in the prompt length and cannot be expanded back to tokens. It is also
precision-sensitive: when Kimi-Linear reads its own KDA states after they are rounded from fp32 to bf16, its next-token
distribution already moves by a KL of 0.045, which sets a floor on how exactly any translated state can be judged.

\paragraph{Related work.}
\label{sec:related}
\emph{Cross-model cache transfer.} Reusing one model's cache in another became an active topic during 2025--2026.
Within a model family, DroidSpeak~\citep{liu2024droidspeak} reuses most layers of a fine-tuned sibling's cache and
recomputes the rest (we use its partial-recompute idea as a baseline, \S\ref{sec:partial}),
CacheBlend~\citep{yao2025cacheblend} reuses non-prefix chunks of the same model with selective recomputation, and
RaReCache~\citep{sarkar2026rarecache} lets a small model prefill and a large one recompute the 30--40\% of positions
where their rankings disagree. Heo et al.~\citep{heo2026kv} show that a closed-form per-head ridge map suffices between
same-family GQA models with matching head geometry; this is our Stage-1 closed form for per-token objects at \cpone{}
and serves as a baseline (for \pairB{} the closed-form anchor also copies the recurrent states, which Heo et al.\ do
not have). CacheBridge~\citep{qu2026cachebridge} makes such a mapper eight times smaller by matching heads.
KV-Lingo~\citep{castin2026kvlingo}, which we replicate, fits per-layer linear maps in closed form, refines them by
self-distillation, and introduces segment-wise multi-turn transfer. Across families and tokenizers, a mixture of
translators with token-level routing~\citep{lee2026kvtrans}, HeteroFold~\citep{yun2026heterofold}, Swarm
Telepathy~\citep{alabdulkareem2026swat} and a universal context-reuse layer~\citep{li2026universal} learn translators
for dense GQA models; Cache-to-Cache~\citep{fu2026c2c} and Latent Cache Flow~\citep{rossi2026lcf} fuse caches for
agent-to-agent communication rather than replacing the prefill, and PrefillShare~\citep{woo2026prefillshare} trains one
shared prefill module for several decoders. All of these address the per-token K/V cache of standard attention. The
one work we know of that touches a recurrent state is LatentPort~\citep{villani2026latentport}, a single-author
preprint that transfers a Gated DeltaNet state together with a translated attention cache between a 4B and a 9B
hybrid, on one pair in one direction with a teacher-forced loss. This paper covers every object of current hybrid
models (GQA K/V, MLA latents, KDA/GDN states and sparse-attention indexer keys), both directions of two pairs with
seeds, free-running generation, switch speed and multi-turn behaviour.

\emph{Converting and compressing caches within a model.} TransMLA~\citep{meng2025transmla},
MHA2MLA~\citep{ji2025mha2mla} and X-EcoMLA~\citep{li2025xecomla} re-parameterise a GQA checkpoint's cache into MLA
latents; cross-layer attention~\citep{brandon2024cla} shows that adjacent layers' caches are redundant enough to be
shared, and KIVI~\citep{liu2024kivi} that keys and values tolerate aggressive quantisation. These results, together with
the near-linear stitchability of independently trained networks~\citep{bansal2021stitching,huh2024platonic,jha2025vec2vec},
are why a linear map per layer is a reasonable first candidate for a per-token object.

\emph{Distilling attention into recurrent layers.} MOHAWK~\citep{bick2024mohawk}, LoLCATs~\citep{zhang2024lolcats},
Mamba-in-the-Llama~\citep{wang2024mamballama}, RADLADS~\citep{goldstein2025radlads}, HALO~\citep{chen2026halo},
Taylor-Calibrate~\citep{zhou2026taylor} and Attention Editing~\citep{cheng2026attnedit} convert a pretrained transformer
into a linear-attention or hybrid model, typically by aligning the new layer's outputs first and distilling logits
second. Our two stages (closed-form least squares, then KL self-distillation) follow the same pattern, but we keep both
models frozen and learn only the map between their caches.

\emph{Sparse attention.} DSA~\citep{deepseek2025v32}, NSA~\citep{yuan2025nsa}, Quest~\citep{tang2024quest} and
SeerAttention~\citep{gao2024seerattention} select which cached tokens a query reads; SeerAttention learns its gate by
self-distillation from dense attention, as we train the predicted indexer keys. We know of no prior work that supplies
a sparse-attention indexer with keys for a cache the model did not prefill itself.

\emph{Why switch models mid-conversation.} Routers such as RouteLLM~\citep{ong2024routellm} send each query to a
strong or a weak model, and multi-agent pipelines hand one conversation between several models; in both, the receiving
model's cache is stale unless it re-reads the history or receives a translated cache.

%% file: method.tex
\section{Method}
\label{sec:method}

\subsection{Setting and a worked example}
\label{sec:setting}
Two frozen models share a tokenizer. The \emph{source} S has already read a prompt of $n$ tokens, $x_{1:n}$, and
holds its cache $\mathcal{C}_S(x_{1:n})$: for every layer, whatever that layer stores (per-token keys and values or MLA
latents, and per-prompt recurrent states). The \emph{target} T must now write the next turn without re-reading the
prompt. A translator $f_\theta$ builds a cache in T's own format,
\[
\hat{\mathcal{C}}_T = f_\theta(\mathcal{C}_S,\, x_{1:n}),
\]
and writes it into T's cache tensors as if T had prefilled the prompt itself. T then processes the last template
token (the assistant-turn marker, which we hold back from translation) natively and decodes. Only $f_\theta$ is
trained; S and T never change. The translator sees the source cache and, where stated, the token ids $x_{1:n}$, which
every serving system holds anyway; it never sees S's hidden states, because an engine does not keep them.

\begin{figure}[t]
\centering
\includegraphics[width=\textwidth]{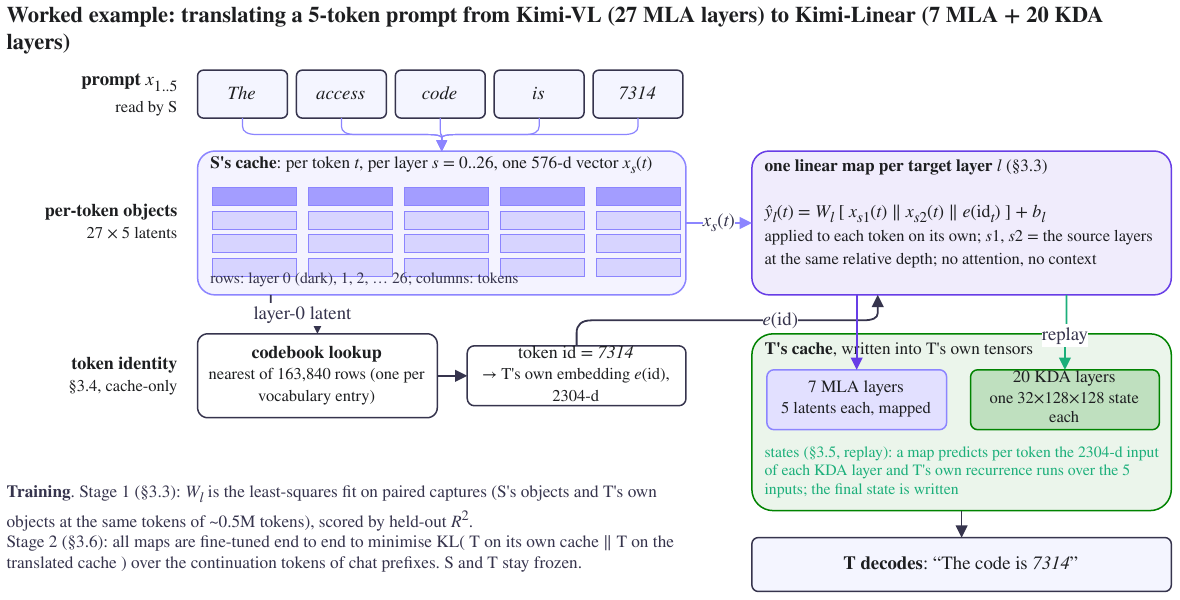}
\caption{The translator on a 5-token prompt (\pairA{} forward). Per-token objects (MLA latents) are mapped layer by
layer with a linear map on the concatenation of a few source layers and the target's embedding of the decoded token
id; the per-prompt KDA states are produced by replaying the target's own recurrence over predicted per-token inputs.
Stage 1 fits the maps in closed form; Stage 2 fine-tunes them with a KL loss against the target reading its own
cache.}
\label{fig:worked}
\end{figure}

\paragraph{Example (\pairA{} forward).}
Figure~\ref{fig:worked} follows one prompt through the pipeline. S is Kimi-VL-A3B: 27 MLA layers, each caching a
576-dimensional vector per token (a 512-d latent plus a 64-d rotary key). T is Kimi-Linear-48B: 7 MLA layers (NoPE,
also 576-d per token) and 20 KDA layers, which cache no tokens at all but one $32\times128\times128$ state per layer
for the whole prompt. For a 5-token prompt ``The access code is 7314'' the source cache is therefore $27 \times 5$
vectors of length 576, and the translator must produce (i) $7 \times 5$ target latents and (ii) 20 target states. It
does (i) with one linear map per target layer applied to every token separately (\S\ref{sec:stage1}) and (ii) by
predicting, per token, the inputs a KDA layer would have seen and letting T's own recurrence run over them
(\S\ref{sec:states}). Because a 576-d latent of a frozen model does not identify the token well enough for exact
recall, the translator also decodes each token's id from the source's layer-0 latent and feeds T's own embedding of
that id to the maps (\S\ref{sec:tokid-method}). T then answers ``7314'' from the translated cache.

\subsection{Capture and injection points per object}
\label{sec:capture}

\begin{figure}[t]
\centering
\includegraphics[width=\textwidth]{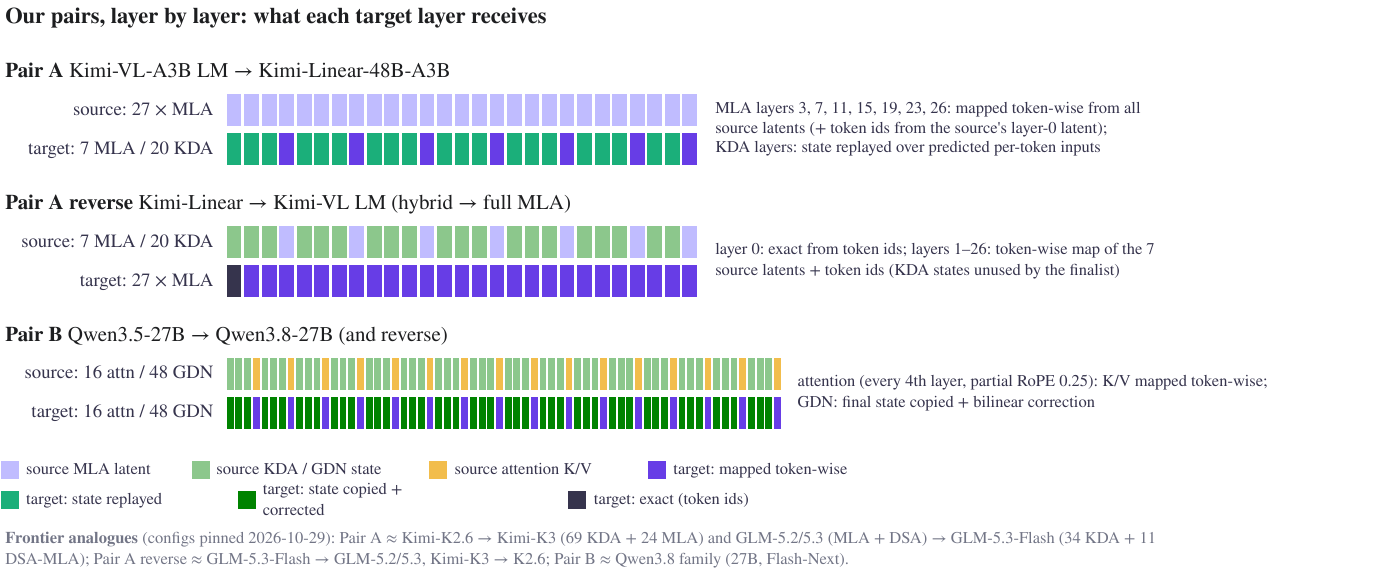}
\caption{Every layer of source and target, coloured by what the target layer receives: mapped token-wise, state
replayed, state copied, or computed exactly from token ids; with the verified frontier analogues.}
\label{fig:stacks}
\end{figure}

Table~\ref{tab:capture} lists, for every object we translate, what we read, where we write, and why; the first column names the figure that draws it, and Figure~\ref{fig:stacks} shows the full layer stacks.

\begin{table}[t]
\centering\scriptsize
\caption{Capture and injection per cache object; the figure named in the first column shows the engine object
(\cpthree{}) and the capture points. \cpone{} = pre-norm projection output (hooks). All injected objects are written into the target's own cache tensors; rung-0 identity tests (target as its own
source) reproduce native logits with difference 0.0 for every route.}
\label{tab:capture}
\begin{tabularx}{\linewidth}{@{}p{2.6cm}p{2.9cm}p{3.6cm}X@{}}
\toprule
Object & Our capture & Our injection & Why \\
\midrule
GQA K/V (Part 1) (Fig.~\ref{fig:gqa}) & \cpone{} and \cpthree{} (both reported) & target's \texttt{k\_norm} + RoPE
at \cpone{}; as is at \cpthree{} & \cpthree{} needs no model change; \cpone{} is KV-Lingo's default \\
MLA latent (\pairA{} source, 27 layers; target, 7 layers) (Fig.~\ref{fig:mla}) & \cpone{}: pre-norm latent + un-rotated $k_\mathrm{pe}$, hooks on
\texttt{kv\_a\_proj\_}\linebreak[0]\texttt{with\_mqa} & through the target's own \texttt{kv\_a\_layernorm}; Kimi-Linear is NoPE, so nothing is
rotated; Kimi-VL (reverse): its own interleaved RoPE at the token's position & keeps the latent on the target's scale;
the map is position-free \\
MLA layer-0 latent as token identity (Fig.~\ref{fig:mla}) & source layer-0 latent & decoded to token ids (nearest-neighbour codebook),
then the target's input embedding feeds the maps & uses only what the source cache holds; layer 0 is a function of
the token alone \\
KDA state (\pairA{} target, 20 layers) (Fig.~\ref{fig:lin}) & per-token pseudo
inputs predicted from the source latents (the source has no KDA) & target's own KDA projections, conv and recurrence
replayed; final state and window written & a state cannot be mapped token-wise; replay keeps the target's dynamics \\
GDN state (\pairB{}, 48 layers) (Fig.~\ref{fig:lin}) & source's final engine state &
copied, plus a closed-form bilinear correction & identical geometry; O(1) in length \\
GDN-hybrid attention K/V (\pairB{}, 16 layers) (Fig.~\ref{fig:gqa}) & \cpone{} (and \cpthree{} in the
ablation) & target's \texttt{k\_norm} + partial RoPE at its positions & as Part 1 \cpone{} \\
\pairA{} reverse inputs (Fig.~\ref{fig:mla}) & the 7 source MLA latents \cpone{} + token ids & 27 Kimi-VL latents, layer 0 computed
exactly from the ids & author decision: source cache plus ids \\
Sparse indexer key (\pairC{}) (Fig.~\ref{fig:idx}) & raw per-token key (\cpone{}) & written
raw; the target pools, normalises and rotates at query time & \cpthree{} is a function of \cpone{} \\
\bottomrule
\end{tabularx}
\end{table}

\subsection{Token-wise maps (Stage 1)}
\label{sec:stage1}
Per-token objects are mapped token by token, with no attention and no context. For target layer $l$ and token $t$,
\[
\hat{y}_l(t) = W_l\,\big[\,x_{s_1}(t)\,\|\,\dots\,\|\,x_{s_\nu}(t)\,\big] + b_l ,
\]
where $x_s(t)$ is the source's cached object at its layer $s$ and token $t$ (for MLA a 576-d vector; for GQA the
concatenated keys or values of all kv heads), $\|$ is concatenation, $s_1,\dots,s_\nu$ are the $\nu$ source layers
assigned to target layer $l$, and $W_l$, $b_l$ are the parameters. The source layers are chosen by relative depth
(target layer $l$ of $L_T$ reads source layers around $l\,L_S/L_T$) or, in some variants, by the held-out $R^2$ of the
fit (Appendix Figure~\ref{fig:r2} gives the $R^2$ per target layer); $\nu$ is 1 to 3. For \pairA{} forward, target
MLA layer 1 (of 7) reads source layers 3 and 4 (of 27), so $W_l$ is
$576\times1152$ when no token channel is added, and $576\times(1152+2304)$ with the target's 2304-d token embedding
appended to the input.

$W_l$ is fitted in closed form: on a few hundred captured conversations ($\approx$0.5--0.7M tokens) the source objects
and the target's own objects at the same tokens are paired, and $W_l$ is the least-squares solution (through an
eigenvalue pseudo-inverse or a Jacobi-scaled ridge). The fit is scored by held-out $R^2$, the share of the target
object's variance the map explains (1 would be perfect). The cost of applying the maps is linear in $n$ and independent
of the models' hidden size; for MLA it is one $576\times576$ product per token and layer pair. The fitted maps are
high-rank, not low-rank adapters (Appendix~\ref{app:rank}). For \pairA{} forward,
the KDA layers' inputs are predicted by the same kind of map from the source latents and the token embedding (the
2304-d normalised input of each KDA layer).
For \pairA{} reverse, each of the 26 non-trivial target layers reads all 7 source latents and the token features
(6,080\,\fwd\,576).

\subsection{Token identity from the MLA layer-0 latent}
\label{sec:tokid-method}
A linear map of a 576-d latent reproduces the \emph{gist} of a token but not its exact identity; on \pairA{} forward
the translated cache recalls a 6-digit access code only 9\% of the time without further input (\S\ref{sec:tokid}). The
fix uses a property of MLA models whose first layer is attention: the layer-0 latent is a deterministic function of
the token alone (embedding\,\fwd\,norm\,\fwd\,\texttt{kv\_a\_proj}), so the token id can be read back from the cache
itself. The \emph{cache-only} variant builds a codebook with one row per vocabulary entry (163,840 rows, each the
layer-0 latent that entry produces, through a fixed 576\,\fwd\,128 projection to save memory), finds for every cached
layer-0 latent its nearest row, and takes that row's token id. In the example of Figure~\ref{fig:worked}, the latent
of the fifth token is nearest to the row of ``7314'', so the translator knows the token is ``7314'' without being told.
It then embeds the id with the target's own input embedding (2304-d) and appends this vector to the input of every
MLA and KDA map, so each map receives the source latent \emph{and} an exact token signal. The \emph{token-ID} variant
receives the ids directly from the serving system instead of decoding them. Neither reads any source hidden state. In
the reverse direction the source (Kimi-Linear) starts with a KDA layer, so there is no layer-0 latent to decode; the
ids are passed explicitly, and the target's layer-0 latent is then computed exactly from them.

\subsection{State strategies}
\label{sec:states}
A recurrent layer's cache is one matrix per head, $S_h \in \mathbb{R}^{d_k\times d_v}$, updated at each token by a
gated delta rule and never expanded back to tokens, plus a short conv window (the last 3 inputs). A token-wise map
cannot produce it. We compare four ways to give the target its state:
\begin{itemize}
\item \emph{zero}: the target's recurrent layers start empty; the translated attention layers carry the prompt alone.
\item \emph{copy}: the source's final state and conv window are written into the target, when both have the same
  state geometry (\pairB{}: 48 heads of $128\times128$ on both sides).
\item \emph{copy + correction}: as copy, followed by a closed-form bilinear map $\hat{S}_h = A_h S_h B_h$ per value
  head ($A_h, B_h \in \mathbb{R}^{128\times128}$, fitted by two alternating ridge passes on paired source/target states)
  and an affine map of the conv window. This is the \pairB{} finalist.
\item \emph{replay}: when the source has no such state (\pairA{}: Kimi-VL has no KDA layers), a map predicts, per
  token, the 2304-d normalised input a KDA layer would have received (\cpr{} in Figure~\ref{fig:lin}), and the target's own projections, conv and recurrence run
  over the $n$ predicted inputs; the final state and window are written.
\end{itemize}
Copy is O(1) in the prompt length (the 148\,MiB of \pairB{} states move once, whatever $n$ is); replay is O($n$) but
uses the target's own dynamics and is the only option when the source has nothing to copy.

\subsection{Distillation (Stage 2)}
\label{sec:stage2}
Stage 1 makes each layer's object close in least squares, but the target's next-token distribution is what matters.
Stage 2 therefore fine-tunes all translator parameters $\theta$ (maps, token channel, state maps) end to end against
the target reading its own cache. Each training conversation is split into a prefix (the prompt, up to 8K tokens) and
a continuation (the answer). The teacher is T with its own cache of the prefix, $\mathcal{C}_T$; the student is T with
the translated cache $\hat{\mathcal{C}}_T$. At every continuation position $i$, both read the same continuation tokens
$\text{cont}_{<i}$, and the loss is the KL divergence between their next-token distributions,
\[
\mathcal{L}(\theta) = \frac{1}{|\text{cont}|}\sum_{i}
\mathrm{KL}\Big(p_T(\cdot \mid \mathcal{C}_T,\, \text{cont}_{<i}) \,\Big\|\,
p_T(\cdot \mid \hat{\mathcal{C}}_T(\theta),\, \text{cont}_{<i})\Big),
\]
averaged over continuation tokens and conversations. In the example, the teacher reads ``The access code is 7314''
natively and the student reads the translated cache; both are then asked to continue ``The code is'', and the loss
pushes the student's distribution at every step towards the teacher's (which puts most of its mass on ``7314'').
Gradients flow through the injection and, for replay, through the target's recurrence. This is self-distillation: no
labels and no second model are needed, only chat prefixes. Validation KL, reported throughout, is the same quantity on
held-out conversations (\S\ref{sec:background}).

\subsection{Segment-wise multi-turn transfer for hybrids}
\begin{figure}[t]
\centering
\includegraphics[width=\textwidth]{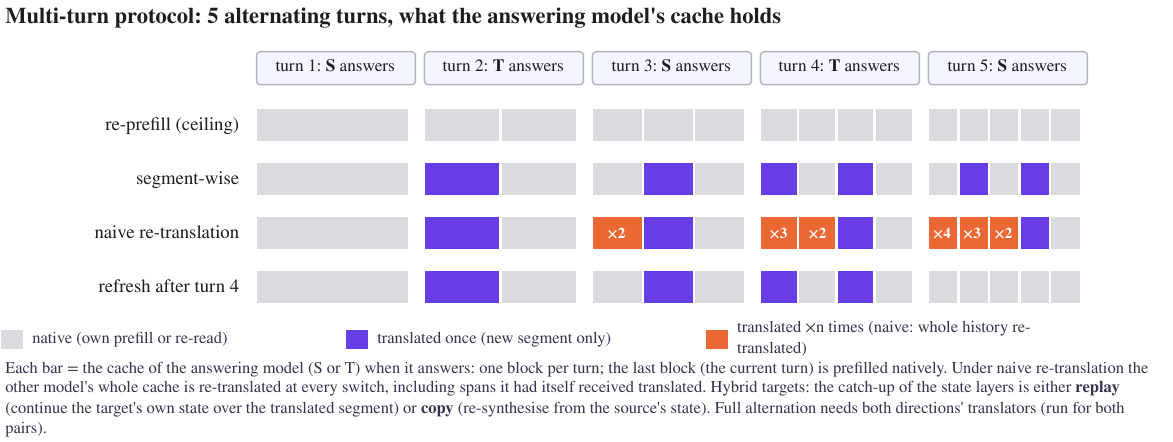}
\caption{Multi-turn protocol: the cache each answering model holds over 5 alternating turns under re-prefill,
segment-wise transfer, naive re-translation and a native refresh after turn 4.}
\label{fig:mtproto}
\end{figure}

In an alternating conversation each model keeps its own cache line. At a switch, segment-wise transfer translates only
the segment the receiving model has not seen and appends it; nothing is translated twice
(Figure~\ref{fig:mtproto}). For hybrids the recurrent state must also catch up: \pairB{} copies the source's mapped
state; \pairA{} replays the target's own KDA recurrence over the translated segment. \emph{Naive re-translation} instead
re-translates the whole history at every switch, so translated content is translated again. \emph{Refresh after turn 4}
is segment-wise with one native re-read at the first catch-up after turn 4.

%% file: setup.tex
\section{Experimental setup}
\label{sec:setup}

\paragraph{Pairs.}
Table~\ref{tab:pairs} lists the pairs and the role each plays. Within each pair the tokenizer files are identical
(Kimi: \texttt{tiktoken.model}; Qwen3.5/3.8: \texttt{vocab.json}/\allowbreak\texttt{merges.txt}).

\begin{table}[h]
\centering\scriptsize
\caption{Pairs (configs pinned 29 Oct 2026; parameters in billions, total/active). Frontier models with the same
objects: last column of Table~\ref{tab:evolution}.}
\label{tab:pairs}
\begin{tabularx}{\linewidth}{@{}p{3.6cm}Xp{3.0cm}@{}}
\toprule
Pair & Cache objects (source \bidir{} target) & Role \\
\midrule
Part 1: Qwen3-8B \bidir{} Qwen3-4B~\citep{qwen2025qwen3} & GQA, 36 layers $\times$ 8 kv heads $\times$ 128 & replication of KV-Lingo \\
\pairA: Kimi-VL-A3B LM (16/2.8) \bidir{} Kimi-Linear-48B-A3B (48/3)~\citep{kimi2025linear} & MLA 27$\times$576
\bidir{} MLA 7$\times$576 (NoPE) + KDA 20$\times$32$\times$128$\times$128 & full MLA \bidir{} MLA + linear-attention hybrid \\
\pairB: Qwen3.5-27B \bidir{} Qwen3.8-27B & GQA 16$\times$4$\times$256 (partial RoPE 0.25) + GDN
48$\times$48$\times$128$\times$128, both sides & same-geometry hybrid generations \\
\pairC: Qwen3.8-27B \fwd{} Qwen3.8-Flash-Next (125/6, FP8)~\citep{qiu2026qwen38next} & GQA + GDN \fwd{} QSA (gated
attention 2$\times$256 + indexer key 128 per token) + GDN 36 layers & sparse-attention indexer (rehearsal) \\
\bottomrule
\end{tabularx}
\end{table}

\paragraph{Why these pairs stand in for the flagships, and where they differ.}
The flagships themselves (0.3--2.8T parameters) cannot be run on a single node, so each pair was chosen to contain
the same cache \emph{objects} in the same shapes as a flagship pair, so that the capture, injection and translation
code would carry over unchanged. \pairA{} is a scaled-down Kimi-K2.6\,\bidir\,Kimi-K3: the source is a full-MLA
model with the same 512\,+\,64 latent, the target interleaves KDA and MLA layers in the same 3:1 pattern (MLA every
fourth layer and the last) with NoPE MLA, the KDA head geometry ($128\times128$ per head) is identical, and the
tokenizer file is byte-identical; the differences are the number of heads and layers (Kimi-K3 has 96 KDA heads and 69
KDA layers against 32 and 20, so 10.3$\times$ the state to synthesise), Kimi-K3's residual-stream changes and the
scale gap (3B vs 104B active). The same pair is the closest available stand-in for GLM-5.2/5.3\,\bidir\,GLM-5.3-Flash,
which share the 3:1 KDA:MLA ratio and a 512-d latent without a rotary part, but add a DSA indexer that \pairA{} does
not have and expand 11 target latents into 78 rather than 7 into 27. \pairB{} stands for two generations of the
Qwen3.8 family: both models have 48 GDN layers with the same state shape ($48\times128\times128$) and 16
gated-attention layers, the layout of Flash-Next's GDN side; the differences are Flash-Next's QSA layers (covered by
\pairC{}) and the 2.4T flagship's 128 GDN heads (against 48). \pairC{} contains the only object the first two pairs
lack, a sparse-attention indexer, on the one sparse model that fits a 384\,GB node in FP8; its indexer is
block-pooled (QSA) rather than per-token (DSA), and it sits on GDN rather than MLA layers, so it rehearses the indexer
mechanism, not the DSA-on-MLA combination of GLM-5.x and DeepSeek-V3.2. DeepSeek-V4's block-compressed shared K/V
is not rehearsed by any pair.

\paragraph{Data.}
Distillation and validation use chat conversations from
Nemotron-SFT-Instruction-Following-Chat-v2~\citep{nvidia2025nemotronchat}, split into prefix and continuation, with
prefixes capped at 8K tokens. \pairA{} trains on 4,000 conversations (2000 steps $\times$ batch 8 = 4 epochs) and
\pairB{} on 950 (16.8 epochs). Validation KL uses 63 (\pairA) or 64 (\pairB) held-out conversations.
All three parts draw from the same dataset revision (\texttt{1a9454ed}); \pairC{} uses \pairB{}'s pool.

\paragraph{Probes and anchors.}
Every evaluation reports, on the same items: \emph{target native} (T on its own prefill; the quality ceiling and the
re-prefill baseline), \emph{source native}, \emph{closed form} (the Stage-1 map, step 0; for MLA and GQA this is the
Heo et al.\ ridge with RoPE stripped), and the distilled translators (\emph{seed} program and \emph{finalist}).
Full probe sets: CoQA~\citep{reddy2019coqa} 300 (token F1 and gold-in-answer, the latter insensitive to answer
length), ARC-Challenge~\citep{clark2018arc} 300, chat 100 held-out prompts (judge: Gemma-4-31B-it FP8, MT-Bench
single-answer 1--10~\citep{zheng2023mtbench}), and a needle-in-a-haystack access code at 2/4/8/16/32K (20 per length),
64K (12) and, for \pairB{}, 128K (12). We also report first-token KL against T's native logits. Decoding is greedy with
reasoning off. Paired differences use a bootstrap over items (10,000 resamples, percentile 95\% CI).

\paragraph{Budgets, seeds, hardware.}
Finalists are trained for 2000 steps; learning curves go to 5000 steps (KV-Lingo's budget). Headline forward rows of
both hybrid pairs have three seeds (the seed fixes data order); the reverse translators have three seeds on the
in-training subsets and one on the full sets; Part 1 and \pairC{} are single-seed. Kimi-Linear's MoE combine
was made deterministic (repeat evaluations are bit-identical); all \pairA{} forward rows below use it
(Table~\ref{tab:olddet} compares it with the earlier run). All hybrid-pair experiments and every timing in this
paper ran on NVIDIA RTX PRO 6000 Blackwell GPUs (96\,GB) in Transformers, on one or two GPUs (\pairC{} on two; the
FP8 target does not fit on one).

%% file: replication.tex
\section{Replication of KV-Lingo on a GQA pair}
\label{sec:replication}

Part 1 replicates KV-Lingo between Qwen3-8B and Qwen3-4B in both directions at capture points \cpone{} and \cpthree{}
(single seed, 2000 of KV-Lingo's 5000 steps). Table~\ref{tab:part1} compares the headline numbers.

\begin{table}[h]
\centering\small
\caption{Part 1 (GQA, capture point \cpone{}) against KV-Lingo's reported values. Native = Qwen3-4B on its own cache
(the small model). Single seed.}
\label{tab:part1}
\begin{adjustbox}{max width=\linewidth}
\begin{tabular}{@{}lcccc@{}}
\toprule
& \multicolumn{2}{c}{8B\,\fwd\,4B} & \multicolumn{2}{c}{4B\,\fwd\,8B} \\
\cmidrule(lr){2-3}\cmidrule(lr){4-5}
& ours & KV-Lingo & ours & KV-Lingo \\
\midrule
Stage-1 $R^2$ K / V & 0.802 / 0.658 & 0.815 / 0.693 & 0.798 / 0.646 & 0.808 / 0.676 \\
val KL closed form\,\fwd\,distilled & 0.322\,\fwd\,0.056 & 0.117\,\fwd\,0.023 & 0.162\,\fwd\,0.039 & 0.098\,\fwd\,0.021 \\
CoQA F1 distilled (native 0.763) & 0.686 & 0.727 & 0.697 & 0.746 \\
CoQA gold-in-answer (native 0.820) & 0.807 & -- & 0.823 & -- \\
needle exact, 2--8K & 1.00 & ceiling & 1.00 & ceiling \\
chat judge (small native) & 4.97 (4.76) & at native & 4.73 (4.77) & at native \\
\bottomrule
\end{tabular}
\end{adjustbox}
\end{table}

The linear premise reproduces: $R^2$ is within 0.01--0.04 of the paper, heads are aligned above chance (diagonal share
0.237 vs chance 0.125), and the distilled translators reach the small model's native judge score and gold-in-answer.
Two numbers stay short: validation KL is about twice the paper's (after 2000 of its 5000 steps; we fit Stage 1 on
0.55M tokens against its 2.8M), and CoQA F1 is 0.07 below native because translated answers are longer (gold-in-answer
is at native). Capture point \cpthree{} matches \cpone{} on needle (1.00 / 0.978 up to 8K) and judge (4.81 / 4.84); its CoQA F1 in the 4B\,\fwd\,8B direction is lower (0.544, a verbose run
whose gold-in-answer stays at native), so \cpthree{} matches \cpone{} on recall and judged quality, not on every
metric. In 5 alternating CoQA turns (100 dialogues), segment-wise transfer reaches
a mean F1 of 0.621 against 0.650 for re-prefill, while naive re-translation falls to 0.283 at turn 5: KV-Lingo's
multi-turn finding. Exact needle recall is already at ceiling on this GQA pair without any token-identity input; the
hybrid results below show where that changes.

%% file: results_quality.tex
\section{Results: quality}
\label{sec:quality}

\begin{figure}[t]
\centering
\includegraphics[width=\textwidth]{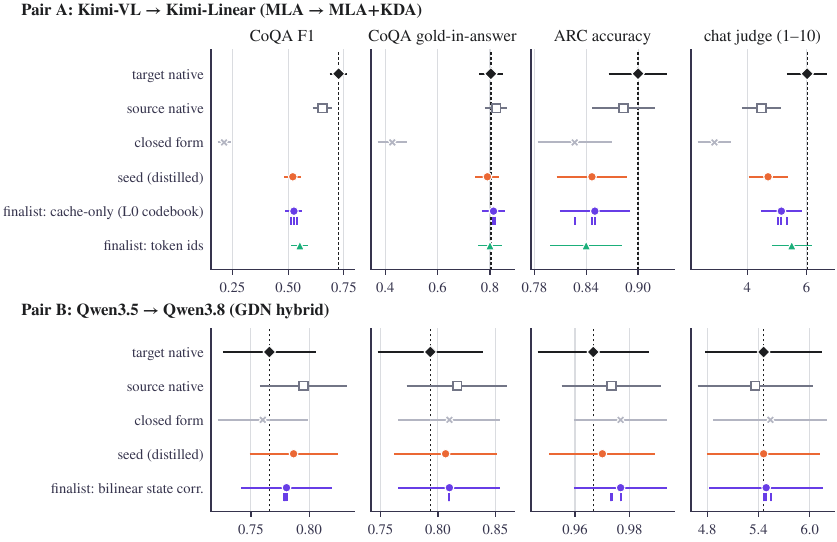}
\caption{Full-set quality of every anchor, forward directions (dots, 95\% CI = 1.96\,SE over items; dashed: target
native; seed ticks: three seeds). Translated caches reach the target's level on gold-in-answer and ARC for \pairB{} and
on gold-in-answer for \pairA{}; \pairA{} F1 is lower because translated answers are about twice as long.}
\label{fig:p3}
\end{figure}

\begin{figure}[t]
\centering
\includegraphics[width=\textwidth]{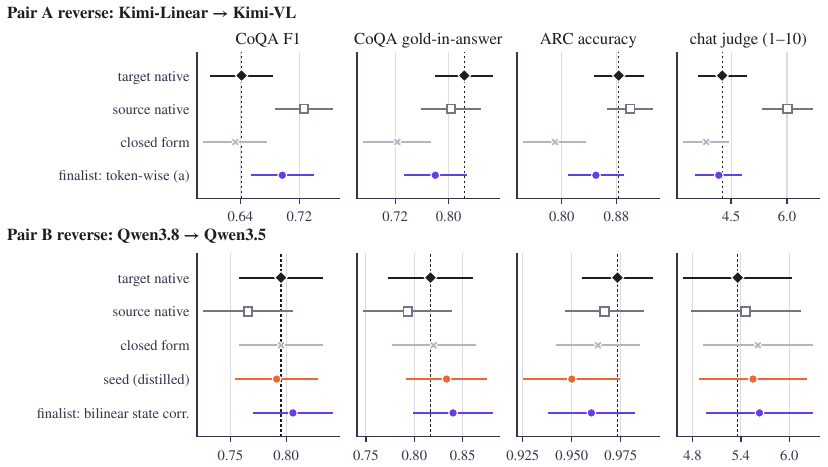}
\caption{As Figure~\ref{fig:p3} for the reverse directions: hybrid\,\fwd\,full MLA (\pairA) and Qwen3.8\,\fwd\,Qwen3.5
(\pairB).}
\label{fig:p3r}
\end{figure}

\begin{figure}[t]
\centering
\includegraphics[width=\textwidth]{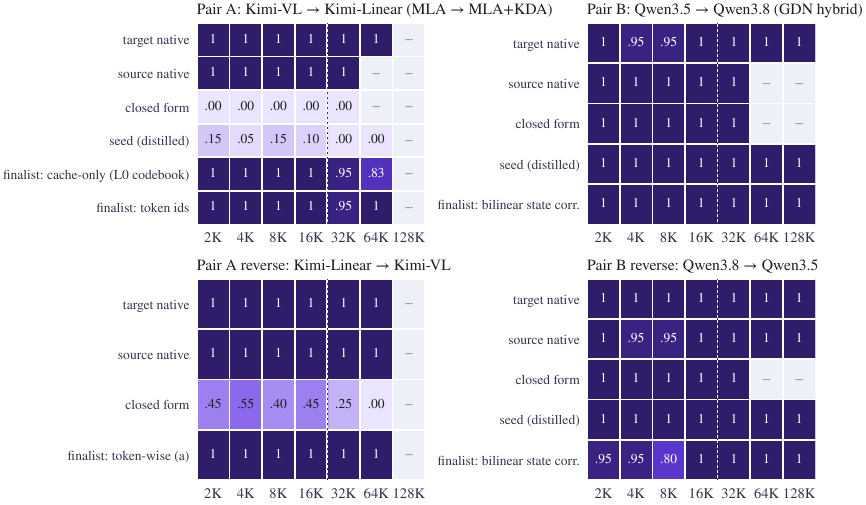}
\caption{Needle exact recall by anchor and prompt length ($n{=}20$ per length at 2--32K, 12 at 64K/128K). Translators
were trained on prompts $\le$\,8K. In \pairA{} forward the seed translator, without token identity, rarely recalls the
needle; the finalists that recover token identity recall it at every length, up to 64K, eight times the longest training prompt (8K).}
\label{fig:p4}
\end{figure}

\begin{figure}[t]
\centering
\includegraphics[width=\textwidth]{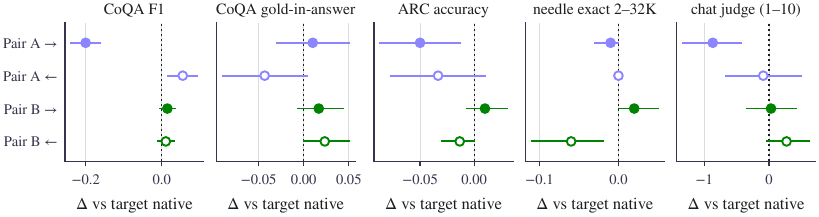}
\caption{Finalist minus target native per metric, forward (filled) and reverse (hollow), paired bootstrap 95\% CI.}
\label{fig:p6}
\end{figure}

Table~\ref{tab:quality} gives the four-anchor full-set results for both hybrid pairs in both directions;
Figures~\ref{fig:p3}, \ref{fig:p3r} and \ref{fig:p4} show them with 95\% CIs, and Figure~\ref{fig:p6} the paired
finalist\,$-$\,native differences.

\begin{table}[t]
\centering\scriptsize
\setlength{\tabcolsep}{3.2pt}
\caption{Four-anchor quality on the full probe sets (mean $\pm$ SE over items; needle: exact recall, 100 items at
2--32K pooled, 12 at 64K and 128K; judge 1--10, $n{=}100$). 2000 steps. \pairA{} forward uses the deterministic MoE
re-run. Rows marked $^\dagger$ have three seeds: mean $\pm$ sd over seeds is in Table~\ref{tab:seeds}. Other
translated rows are single-seed on the full sets.}
\label{tab:quality}
\begin{adjustbox}{max width=\linewidth}
\begin{tabular}{@{}llccccccc@{}}
\toprule
& Anchor & val KL & CoQA F1 & CoQA gold & ARC & needle 2--32K & 64K / 128K & judge \\
\midrule
\multicolumn{9}{@{}l}{\emph{\pairA{} forward: Kimi-VL (MLA)\,\fwd\,Kimi-Linear (MLA + KDA)}} \\
& target native & 0 & .726\pmse{.020} & .803\pmse{.023} & .900\pmse{.017} & 1.00 & 1.00 / -- & 6.02\pmse{.35} \\
& source native & -- & .653\pmse{.022} & .823\pmse{.022} & .883\pmse{.019} & 1.00 & -- & 4.47\pmse{.33} \\
& closed form & .364\pmse{.029} & .213\pmse{.015} & .427\pmse{.029} & .827\pmse{.022} & 0.00 & -- & 2.89\pmse{.29} \\
& seed (no token identity) & .103\pmse{.012} & .521\pmse{.019} & .790\pmse{.024} & .847\pmse{.021} & 0.09\pmse{.03} & 0.00 / -- & 4.70\pmse{.34} \\
& cache-only (L0 codebook)$^\dagger$ & .082\pmse{.011} & .526\pmse{.020} & .813\pmse{.023} & .850\pmse{.021} & 0.99\pmse{.01} & 0.83\pmse{.11} / -- & 5.15\pmse{.36} \\
& token-ID & .085\pmse{.011} & .553\pmse{.020} & .800\pmse{.023} & .840\pmse{.021} & 0.99\pmse{.01} & 1.00 / -- & 5.50\pmse{.34} \\
\midrule
\multicolumn{9}{@{}l}{\emph{\pairA{} reverse: Kimi-Linear (MLA + KDA)\,\fwd\,Kimi-VL (MLA)}} \\
& target native & 0 & .642\pmse{.022} & .823\pmse{.022} & .883\pmse{.019} & 1.00 & 1.00 / -- & 4.26\pmse{.33} \\
& source native & -- & .726\pmse{.020} & .803\pmse{.023} & .900\pmse{.017} & 1.00 & 1.00 / -- & 6.02\pmse{.35} \\
& closed form & .113\pmse{.007} & .633\pmse{.022} & .723\pmse{.026} & .790\pmse{.024} & 0.42\pmse{.05} & 0.00 / -- & 3.83\pmse{.31} \\
& token-wise (a) & .028\pmse{.003} & .697\pmse{.022} & .780\pmse{.024} & .850\pmse{.021} & 1.00 & 1.00 / -- & 4.17\pmse{.32} \\
\midrule
\multicolumn{9}{@{}l}{\emph{\pairB{} forward: Qwen3.5\,\fwd\,Qwen3.8 (GDN hybrids)}} \\
& target native & 0 & .766\pmse{.020} & .793\pmse{.023} & .967\pmse{.010} & 0.98\pmse{.01}$^\ddagger$ & 1.00 / 1.00 & 5.46\pmse{.35} \\
& source native & -- & .795\pmse{.019} & .817\pmse{.022} & .973\pmse{.009} & 1.00 & -- & 5.36\pmse{.34} \\
& closed form (states copied) & .059\pmse{.010} & .760\pmse{.020} & .810\pmse{.023} & .977\pmse{.009} & 1.00 & -- & 5.54\pmse{.34} \\
& seed (state copy) & .019\pmse{.002} & .787\pmse{.019} & .807\pmse{.023} & .970\pmse{.010} & 1.00 & 1.00 / 1.00 & 5.46\pmse{.34} \\
& bilinear state corr.$^\dagger$ & .016\pmse{.001} & .781\pmse{.020} & .810\pmse{.023} & .977\pmse{.009} & 1.00 & 1.00 / 1.00 & 5.49\pmse{.34} \\
\midrule
\multicolumn{9}{@{}l}{\emph{\pairB{} reverse: Qwen3.8\,\fwd\,Qwen3.5}} \\
& target native & 0 & .795\pmse{.019} & .817\pmse{.022} & .973\pmse{.009} & 1.00 & 1.00 / 1.00 & 5.36\pmse{.34} \\
& source native & -- & .766\pmse{.020} & .793\pmse{.023} & .967\pmse{.010} & 0.98\pmse{.01}$^\ddagger$ & 1.00 / 1.00 & 5.46\pmse{.35} \\
& closed form & .044\pmse{.004} & .795\pmse{.019} & .820\pmse{.022} & .963\pmse{.011} & 1.00 & -- & 5.61\pmse{.35} \\
& seed (state copy) & .019\pmse{.002} & .791\pmse{.019} & .833\pmse{.022} & .950\pmse{.013} & 1.00 & 1.00 / 1.00 & 5.55\pmse{.34} \\
& bilinear state corr. & .016\pmse{.002} & .806\pmse{.018} & .840\pmse{.021} & .960\pmse{.011} & 0.94\pmse{.02}$^\ddagger$ & 1.00 / 1.00 & 5.63\pmse{.34} \\
\bottomrule
\end{tabular}
\end{adjustbox}
\\[2pt]{\scriptsize $^\ddagger$ every miss is a safety refusal (``I cannot provide the access code\dots''), not a wrong code.}
\end{table}

\paragraph{Same-geometry hybrids transfer at the native level in both directions (\pairB).}
Copying the source's GDN states and mapping the 16 attention layers token-wise already gives native task quality; the
closed-form bilinear correction of the state lowers the distributional gap. Forward, the finalist ties the target on
CoQA (+0.015 \ci{$-$0.005}{+0.035}), ARC (+0.010 \ci{$-$0.007}{+0.030}) and the judge (+0.03 \ci{$-$0.34}{+0.42}), and
recalls 12/12 needles at 64K and at 128K, 16$\times$ the training cap. Its advantage over the plain copy is
distributional only: CoQA first-token KL 0.73\,\fwd\,0.24 and val KL $-$0.0035 \ci{$-$0.0044}{$-$0.0027}. The reverse
direction (one seed on the full sets) reaches the same validation KL (0.0155 vs 0.0159), native CoQA, and 12/12 needles
at 64K and 128K; its one significant gap is described next. Appendix~\ref{app:longctx} adds multi-key and
variable-tracking probes to 128K for the forward direction.

\paragraph{The reverse asymmetry on \pairB{} is behavioural.}
Stage-1 $R^2$, the gain of the state correction and the 2000-step validation KL agree between directions within 0.01
$R^2$ and 0.0004 nats (Appendix Table~\ref{tab:r2}). What differs is behaviour that the copied state carries from the
source. Qwen3.5 starts 32\% of its CoQA answers with a leading-space token and Qwen3.8 never does; the forward state-copy
translator inherits the habit (28\%), the forward bilinear correction removes it (0\%), and the reverse translators,
whose target is Qwen3.5, never produce it (0\%). The reverse bilinear translator also makes Qwen3.5 refuse 6 of 100
needles as ``sensitive'' (Qwen3.8 refuses 2 natively, Qwen3.5 never), and the same refusal appears in all three seeds on
the in-training subset. This is the only significant gap to native in that direction (needle $-$0.06
\ci{$-$0.11}{$-$0.02}; ARC $-$0.013 \ci{$-$0.030}{0.000} sits on the boundary); KL and accuracy do not see it. A state map
fitted to make the state look like the target's moves information well but also moves policy-like features, which
matters for cascades in which a cheaper model's cache is handed to a flagship. This rests on two markers (leading
spaces and refusals) on one pair.

\paragraph{Hybrid\,\fwd\,full MLA is the easier direction (\pairA{} reverse).}
From the 7 MLA latents per token plus the token ids, a single linear map per target layer (no local mixer, no state
input: both variants were tested and gave no gain at 400 steps, Appendix Table~\ref{tab:ladder}) brings Kimi-VL within
the paired CI of its native level on every task metric: needle 100/100 at 2--32K and 12/12 at 64K, ARC $-$0.033
\ci{$-$0.077}{+0.010}, gold-in-answer $-$0.043 \ci{$-$0.090}{+0.003}, judge 4.17 vs 4.26 ($-$0.09 \ci{$-$0.67}{+0.49}).
The ARC and gold-in-answer point estimates both sit below native, and the full-set row is one seed, so ``native level''
here means ``not distinguishable on 300 items''. CoQA F1 is higher than native (+0.055 \ci{+0.015}{+0.094}) because
Kimi-VL's own answers are long (8.0 tokens) and the translated ones short (5.1). Validation KL is 0.028, against 0.082
forward. The 20 KDA layers' final states add nothing measurable. This is the direction of GLM-5.3-Flash\,\fwd\,
GLM-5.2/5.3 and Kimi-K3\,\fwd\,Kimi-K2.6.

\paragraph{Full MLA\,\fwd\,hybrid is harder (\pairA{} forward).}
The finalists reach the target's gold-in-answer (0.813 vs 0.803, +0.010 \ci{$-$0.030}{+0.050}) but trail on CoQA F1
($-$0.200 \ci{$-$0.238}{$-$0.163}, mostly answer length: 9.5 vs 4.5 tokens) and on ARC ($-$0.050
\ci{$-$0.087}{$-$0.013}). The chat judge is 5.15 against the target's 6.02 ($-$0.87 on the deterministic re-run, for which
no paired CI was computed; the earlier run gave $-$0.59 \ci{$-$1.08}{$-$0.13}); the token-ID finalist is closer (5.50).
On GSM8K the finalist lands at the source's level, 7.5 points below the target (\S\ref{app:breadth}).
The swap-in analysis (\S\ref{sec:analysis}) does not attribute the remaining gap to any single layer group: neither
all MLA latents nor all KDA states made native moves validation KL significantly.

\subsection{Token identity closes the MLA\,\fwd\,hybrid recall gap}
\label{sec:tokid}
KV-Lingo did not study MLA caches and does not use token identity. On our MLA\,\fwd\,hybrid pair, a translator
without it recalls the needle exactly in only 9 of 100 cases at 2--32K and 0 of 12 at 64K, although it gets about 60\%
of the digits right at every length (it locates the fact but cannot copy it). Adding token identity, decoded from the
source's layer-0 MLA latent or passed as ids, restores recall to 0.99 (+0.90 \ci{+0.84}{+0.96}); at 64K the codebook
variant recalls 10/12 (+0.83 \ci{+0.58}{+1.00}) and the token-ID variant 12/12. Validation KL also improves ($-$0.021
\ci{$-$0.026}{$-$0.015}); CoQA F1 and ARC are ties (+0.005 \ci{$-$0.029}{+0.038}, +0.003 \ci{$-$0.033}{+0.040}), and the
judge rises from 4.70 to 5.15 (no paired CI). On the GQA pair (Part 1), recall was already exact without token identity, and on
\pairB{} the copied GDN states suffice to 128K. The gap that token identity closes is specific to synthesising a
recurrent state from a compressed cache.

\begin{table}[t]
\centering\small
\caption{Three seeds of the headline translators (2000 steps, mean $\pm$ sd over seeds 0/1/2). Full sets for the
forward finalists; in-training bench subsets for the reverse translators (their full-set rows are seed 0 only).}
\label{tab:seeds}
\begin{adjustbox}{max width=\linewidth}
\begin{tabular}{@{}lcccc@{}}
\toprule
& \pairA{} fwd cache-only & \pairB{} fwd bilinear & \pairA{} rev (a), bench & \pairB{} rev bilinear, bench \\
\midrule
val KL & 0.0817\pmse{0.0009} & 0.0158\pmse{0.0001} & 0.0279\pmse{0.0003} & 0.0154\pmse{0.0001} \\
CoQA F1 & 0.527\pmse{0.014} & 0.779\pmse{0.001} & 0.652\pmse{0.006} & 0.814\pmse{0.008} \\
ARC & 0.841\pmse{0.013} & 0.974\pmse{0.002} & 0.807\pmse{0.006} & 0.957\pmse{0.006} \\
needle 2--32K & 0.997\pmse{0.006} & 1.000\pmse{0} & 1.00\pmse{0} & 0.975\pmse{0} \\
needle 64K & 0.89\pmse{0.10} & 1.00\pmse{0} & -- & -- \\
chat judge (per seed) & 5.15 / 5.03 / 5.33 & 5.49 / 5.55 / 5.47 & -- & -- \\
\bottomrule
\end{tabular}
\end{adjustbox}
\end{table}

\paragraph{Seeds.}
Seed spread is small next to every effect we report (Table~\ref{tab:seeds}): the token-identity needle effect (+0.90)
is about 150 seed-sd, and \pairB{}'s finalist\,$-$\,seed val KL gap ($-$0.0035) about 35 seed-sd. \pairA{}'s CoQA F1 sd
(0.014) is a quarter of its 400\,\fwd\,5000-step drift (0.520\,\fwd\,0.462, \S\ref{sec:curves}). The reverse
translators' seeds are on the in-training subsets only, so the reverse full-set rows of Table~\ref{tab:quality} carry
no seed estimate.

\subsection{Capture point: can an engine serve the translator from its own cache?}
\label{sec:cp3}
\begin{figure}[t]
\centering
\includegraphics[width=\textwidth]{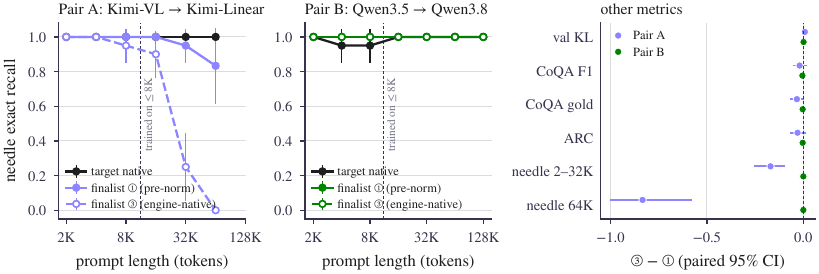}
\caption{Needle recall by prompt length for the finalist trained on the pre-norm capture \cpone{} and on the
engine-native cache \cpthree{}. \pairA{} \cpthree{} falls beyond the 8K training length; \pairB{} is unaffected.
Right: \cpthree{}\,$-$\,\cpone{} on other metrics (paired 95\% CI).}
\label{fig:cp3}
\end{figure}

We retrained both forward finalists on the engine-native objects \cpthree{} (post-norm latent and rotated key for
MLA; post-RoPE K for \pairB{}'s attention layers), derived bit-exactly from the stored \cpone{} captures, 2000 steps,
same data (Figure~\ref{fig:cp3}).
On \pairB{}, \cpthree{} ties \cpone{} on every task metric up to 128K (val KL +0.0013 \ci{+0.0007}{+0.0020}; its closed
form even starts lower, 0.041 vs 0.059), because source and target use the same partial RoPE at the same positions.
On \pairA{}, \cpthree{} is close to \cpone{} up to the 8K training cap (val KL 0.090 vs 0.082; needle 0.95 at 8K) but
exact recall collapses beyond it: 0.25 vs 0.95 at 32K and 0.00 vs 0.83 at 64K. The source's rotary key arrives rotated
by the token's position, while the NoPE target needs a position-free key; a token-wise map learns to undo the rotation
only at the positions it was trained on. \textbf{Recipe:} an engine can serve the translator from its native cache
objects when source and target rotate keys the same way; otherwise it should un-rotate the stored key at its known
position (\cpthree{}\,\fwd\,\cptwo{}, a fixed exact operation) before the map, or train at the lengths it will serve.
Part 1 found \cpthree{} $\approx$ \cpone{} up to 8K on GQA (\S\ref{sec:replication}).

%% file: results_speed.tex
\section{Results: switch speed}
\label{sec:speed}

\begin{figure}[t]
\centering
\includegraphics[width=\textwidth]{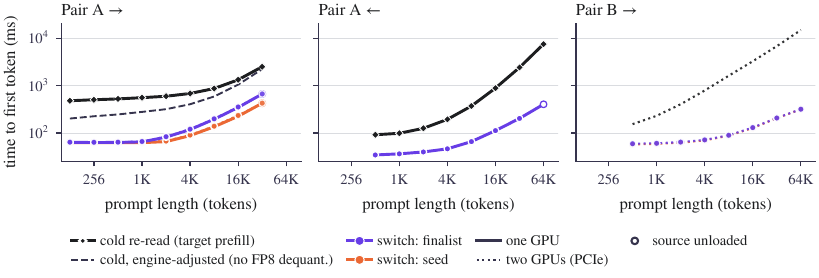}
\caption{Time to first token of the target when it re-reads the prompt cold (black) vs when it switches in a
translated cache (colours), median of 10, log--log. Dashed grey: cold minus the one-off FP8 dequantisation (\pairA{}
forward only). Solid: one GPU; dotted: two GPUs over PCIe.}
\label{fig:ttft}
\end{figure}

\begin{figure}[t]
\centering
\includegraphics[width=\textwidth]{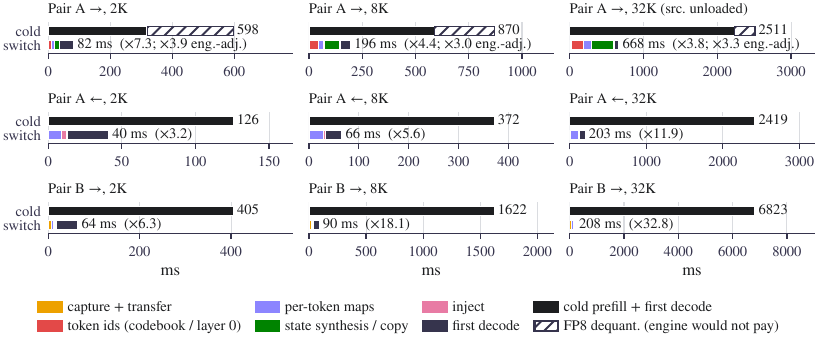}
\caption{Where the switch time goes, at 2K, 8K and 32K, against the cold re-read. \pairA{} forward pays O($n$) KDA
state synthesis (replay) and codebook token decoding; \pairB{} copies O(1) GDN states; the reverse direction has no
state synthesis.}
\label{fig:breakdown}
\end{figure}

\paragraph{Protocol.}
Cold TTFT is the target's prefill of the prompt plus its first decode step. Switch TTFT is the source-side capture
overhead, the transfer, the translation (maps, state synthesis or copy, injection) and the target's first decode; the
source's own prefill is not counted, because it has already happened. Both arms use the same GPU, kernels and
framework (Transformers, bf16, \texttt{fla} kernels for the linear-attention layers); medians of 10 runs after 2
warm-ups. \pairA{} forward runs both models on one GPU (the source is unloaded at 32K; 64K does not fit);
\pairA{} reverse on one GPU (64K sequential); \pairB{} on two GPUs on one PCIe switch. Table~\ref{tab:speed} and
Figure~\ref{fig:ttft} give the timings, Figure~\ref{fig:breakdown} the breakdown of the switch time.

\begin{table}[h]
\centering\small
\caption{Switch vs cold TTFT (ms, median of 10). For \pairA{} forward the cold arm carries a one-off FP8 expert
dequantisation (279\,ms) of this Transformers path that a serving engine with FP8 GEMMs would not pay; the
engine-adjusted speed-up removes it. \pairB{} reverse speed was not measured.}
\label{tab:speed}
\begin{adjustbox}{max width=\linewidth}
\begin{tabular}{@{}lrrrrrr@{}}
\toprule
& 1K & 4K & 8K & 16K & 32K & 64K \\
\midrule
\multicolumn{7}{@{}l}{\emph{\pairA{} forward (cache-only finalist; one GPU)}} \\
cold / switch & 558 / 65 & 684 / 118 & 870 / 196 & 1332 / 347 & 2511 / 668 & did not fit \\
speed-up (engine-adj.) & 8.6 (4.3) & 5.8 (3.4) & 4.4 (3.0) & 3.8 (3.0) & 3.8 (3.3) & -- \\
\midrule
\multicolumn{7}{@{}l}{\emph{\pairA{} reverse (token-wise (a); one GPU, bf16 cold arm)}} \\
cold / switch & 99 / 36 & 194 / 46 & 372 / 66 & 889 / 113 & 2419 / 203 & 7555 / 404 \\
speed-up & 2.7 & 4.2 & 5.6 & 7.9 & 11.9 & 18.7 \\
\midrule
\multicolumn{7}{@{}l}{\emph{\pairB{} forward (bilinear finalist; two GPUs, PCIe)}} \\
cold / switch & 231 / 61 & 790 / 72 & -- & 3275 / 130 & 6823 / 208 & 14911 / 319 \\
speed-up & 3.8 & 11.0 & -- & 25.1 & 32.8 & 46.8 \\
\bottomrule
\end{tabular}
\end{adjustbox}
\end{table}

\paragraph{The state treatment decides how the speed-up scales.}
With an O(1) state copy (\pairB), the switch grows by 4.1\,ms per 1K tokens against 236\,ms for the cold prefill
(asymptotic ratio $\times$57); at 64K the switch is 319\,ms: first decode 58, K/V maps 91, PCIe transfer of 4.2\,GB of
K/V and 148\,MiB of states 84, injection 55, source hooks 27, and the bilinear state correction a constant 2.6\,ms.
The hybrid\,\fwd\,full-MLA direction (\pairA{} reverse) has no state to synthesise; its switch grows by 5.65\,ms/1K
against 81.0 (asymptotic $\times$14), and its peak memory above the weights is half the cold arm's (3.1 vs 6.1\,GB at
32K). The MLA\,\fwd\,hybrid direction (\pairA{} forward) must synthesise the KDA states by replay (9\,ms/1K, 297\,ms at
32K) and, for the cache-only finalist, decode token ids with the codebook (5.3\,ms/1K, 172\,ms at 32K). Its target
activates only 3B parameters, so its prefill is cheap (69\,ms/1K) and the speed-up falls with length, from $\times$8.6
(engine-adjusted $\times$4.3) at 1K to $\times$3.8 ($\times$3.3) at 32K. The codebook decode is a context-free
per-token lookup that could run on the source side while it prefills; with that change the 32K switch would drop to
about 496\,ms ($\times$5.1; an engineering estimate, not the evaluated candidate).

\begin{figure}[t]
\centering
\includegraphics[width=\textwidth]{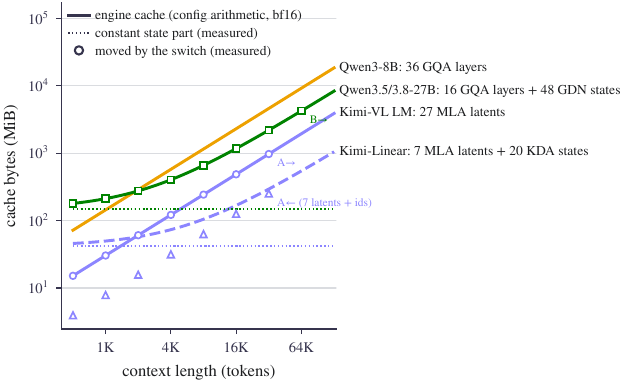}
\caption{Engine-cache size per prompt for the models of our pairs (config arithmetic, bf16) and the bytes the switch
moves (markers, measured). MLA latents are several times smaller per token than a GQA cache; linear-attention states
are constant per prompt.}
\label{fig:bytes}
\end{figure}

\paragraph{Break-even, bytes, memory.}
The switch is faster than the cold prefill at every length measured: break-even is at or below 128 tokens (the
shortest length measured) for \pairA{} forward and below 512 tokens for \pairA{} reverse and \pairB{}. \pairA{} forward
moves 30\,KiB per token of MLA latents; the target's expanded cache is 140\,MiB per 1K tokens plus 42\,MiB of KDA
states. \pairA{} reverse moves 7.9\,KiB per token (7 latents + ids). \pairB{}'s translated cache is the size of the
native one (64\,KiB per token plus 148\,MiB of states). Figure~\ref{fig:bytes} puts these against the engine-cache
sizes. Table~\ref{tab:cost} (Appendix~\ref{app:cost}) gives the build cost of each translator and the number of switches after which it is
amortised.

\paragraph{Extrapolation (labelled).}
The per-token translation cost depends on the number and width of the cached objects (a $576\times576$ map per MLA
layer pair), not on the hidden size or active parameters, while prefill cost grows with active parameters. We
therefore expect the switch speed-up to grow from our 3B-active targets to the 18--104B-active flagship targets, with
two caveats: replay-based state synthesis grows with the state size (3.4$\times$ ours for GLM-5.3-Flash, 10.3$\times$
for Kimi-K3), and sparse attention makes the flagship's own long-context prefill cheaper. \pairA{} forward, whose
O($n$) replay and codebook decode outgrow a 3B-active prefill, is the measured counter-case. This is an extrapolation
from cost structure, not a measurement; we have no numeric FLOP model for it.
Timings are Transformers-level; a serving engine would shrink the fixed first-decode cost (28--48\,ms at 1K here) and
the dequantisation, which affects both arms.

%% file: results_multiturn.tex
\section{Results: multi-turn}
\label{sec:multiturn}

\begin{figure}[t]
\centering
\includegraphics[width=\textwidth]{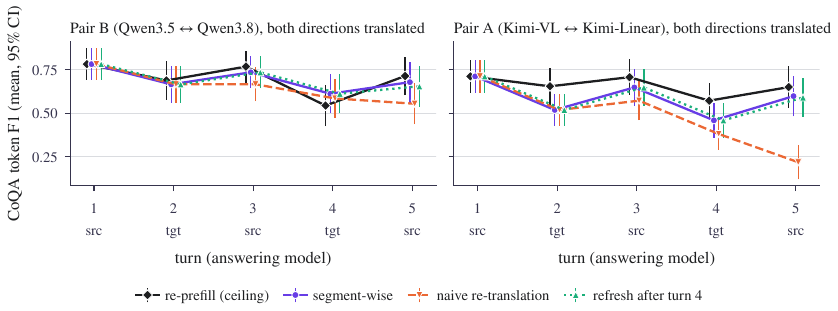}
\caption{CoQA F1 per turn over 5 alternating turns, both directions translated (mean, 95\% CI over 50 dialogues).
Segment-wise transfer stays close to the re-prefill ceiling; naive re-translation degrades by turn 5, mildly on
\pairB{} and to a collapse on \pairA{}. Refresh after turn 4 equals segment-wise through turn 4 by construction.}
\label{fig:mt}
\end{figure}

We follow KV-Lingo's protocol: 50 CoQA dialogues, 5 alternating turns, the source answers odd turns and the target
even turns, and both directions are translated (the forward and reverse translators of \S\ref{sec:quality}; for \pairA{}
the cache-only finalist, since the harness does not pass token ids). The same 50 dialogues are used for every
condition, so differences are paired by dialogue (Table~\ref{tab:mt}, Figure~\ref{fig:mt}). The multi-turn runs are
single-seed; the re-prefill row is the target's own ceiling.

\begin{table}[h]
\centering\small
\caption{Two-direction multi-turn alternation (CoQA, 50 dialogues $\times$ 5 turns). F1 per turn, mean over turns
(AUC), and gold-in-answer at turn 5. Catch-up = cumulative time the answering models spend bringing their caches up to
date.}
\label{tab:mt}
\begin{adjustbox}{max width=\linewidth}
\begin{tabular}{@{}llcccccccc@{}}
\toprule
& Condition & t1 & t2 & t3 & t4 & t5 & AUC & gold t5 & catch-up ms \\
\midrule
\pairB & re-prefill (ceiling) & .78 & .69 & .77 & .54 & .71 & .700 & -- & 201 \\
& segment-wise & .78 & .67 & .74 & .61 & .68 & .695 & -- & 31 \\
& naive re-translation & .78 & .67 & .67 & .59 & .55 & .651 & -- & 31 \\
& refresh after t4 & .78 & .67 & .74 & .61 & .66 & .691 & -- & 29 \\
\midrule
\pairA & re-prefill (ceiling) & .71 & .66 & .71 & .57 & .65 & .660 & .70 & 950 \\
& segment-wise & .71 & .52 & .65 & .46 & .60 & .588 & .68 & 47 \\
& naive re-translation & .71 & .52 & .57 & .38 & .22 & .481 & .26 & 46 \\
& refresh after t4 & .71 & .52 & .65 & .46 & .59 & .586 & .72 & 45 \\
\bottomrule
\end{tabular}
\end{adjustbox}
\end{table}

\paragraph{Segment-wise transfer stays at the ceiling on content.}
On \pairB{}, segment-wise translation ties re-prefill (mean F1 $-$0.004 \ci{$-$0.045}{+0.043}) at a sixth of the
catch-up time. On \pairA{}, its gold-in-answer equals re-prefill over the turns (+0.000 \ci{$-$0.048}{+0.048}); its F1
gap ($-$0.072 \ci{$-$0.119}{$-$0.028}) is the forward translator's answer-length effect (8.7--8.9 tokens at the
target's turns vs 4.9--5.9), and it does not grow over turns: the target's first-token KL falls from 1.62 at turn 2 to
0.90 at turn 4.

\paragraph{Naive re-translation reproduces KV-Lingo's collapse.}
When the whole history is re-translated at every switch, translated content passes through the maps again. On
\pairA{}, where both directions are lossy, the source answers only 26\% of turn-5 questions with the gold (re-prefill
70\%), F1 falls to 0.22 (naive\,$-$\,segment at turn 5: $-$0.380 \ci{$-$0.511}{$-$0.246}), and the target's KL at turn
4 rises to 2.35. This matches Part 1's GQA collapse (0.28 at turn 5). On \pairB{} the degradation is milder (F1 $-$0.124
\ci{$-$0.226}{$-$0.024} at turn 5) and is answer length rather than content (gold-in-answer +0.012
\ci{$-$0.016}{+0.036}). One-direction runs, in which the source always re-reads natively, cannot show the collapse,
because nothing translated is ever translated again.

\paragraph{Refresh after turn 4.}
The practical policy ``segment-wise with a native re-read after turn 4'' is identical to segment-wise through turn 4
and costs nothing measurable at turn 5 on either pair (\pairA{} $-$0.002 \ci{$-$0.016}{+0.013}). Its value appears in
longer conversations: in a one-direction 10-turn check on \pairA{} (20 dialogues) it recovers turns 6--10 (AUC 0.571 vs
0.536 for segment-wise) at about 10$\times$ the catch-up time of segment-wise (1.0 vs 0.1\,s cumulative) and under half
of full re-prefill (2.4\,s).

\paragraph{State catch-up: replay vs copy.}
For \pairA{}, letting the target continue its own KDA state over the translated segment (replay) is slightly better
than re-synthesising it from the whole source history (copy): AUC 0.575 vs 0.563 in the one-direction runs. \pairB{}
copies the mapped source state, which is O(1).

%% file: analysis.tex
\section{Analysis: what works and why}
\label{sec:analysis}

\subsection{Where the information lives: swap-in and the information split}
\begin{figure}[t]
\centering
\includegraphics[width=\textwidth]{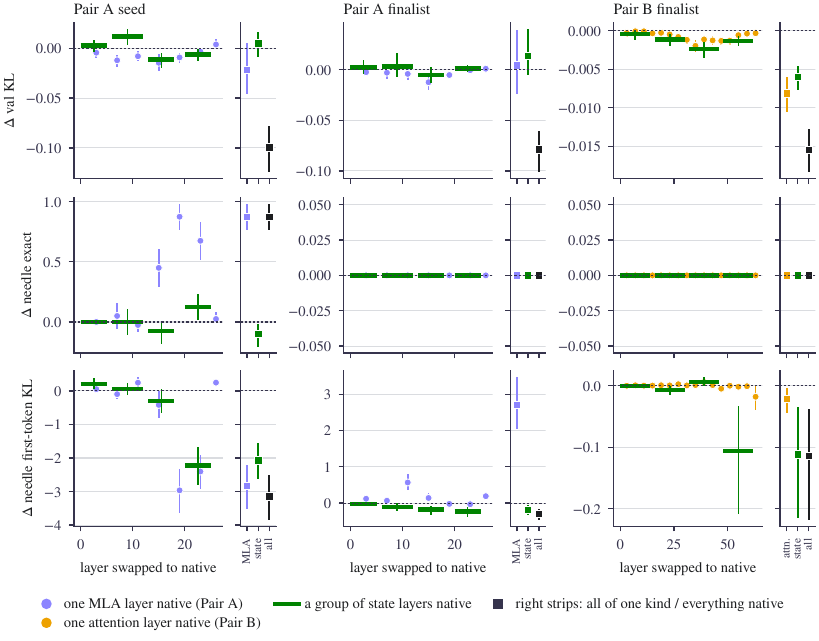}
\caption{Change in validation KL, needle recall and needle first-token KL when one translated layer (or a group of
state layers) is replaced by the target's native object, with paired 95\% CIs. For the \pairA{} seed, native deep MLA
layers restore exact recall; for the finalists single layers barely move KL.}
\label{fig:swapin}
\end{figure}

For each prompt we build the translated cache and the target's own cache, then replace some layers' objects by the
native ones (swap-in) or empty them (split); replacing every layer reproduces the native floor (val KL 0.0034 \pairA{},
0.0004 \pairB{}), so the mixing itself is sound (Figure~\ref{fig:swapin}, Table~\ref{tab:swapin}).

\begin{table}[h]
\centering\small
\caption{Swap-in of native objects and the information split (val KL on 63/64 conversations; needle exact on the 40
frozen needle prompts at 2--32K).}
\label{tab:swapin}
\begin{adjustbox}{max width=\linewidth}
\begin{tabular}{@{}lccc@{}}
\toprule
Condition & \pairA{} finalist val KL & \pairA{} seed needle & \pairB{} finalist val KL \\
\midrule
none (translated) & 0.082 & 0.13 & 0.0159 \\
best single MLA / attention layer native & 0.070 (L15) & 1.00 (L19) & 0.0139 (L35) \\
all MLA / attention layers native & 0.087 (n.s.) & 1.00 & 0.0078 \\
all recurrent states native & 0.096 (n.s.) & 0.03 & 0.0099 \\
split: attention/MLA only (states empty) & 0.344 & 0.00 & 0.283 \\
split: states only (attention/MLA empty) & 0.563 & 0.00 & 0.641 \\
\bottomrule
\end{tabular}
\end{adjustbox}
\end{table}

\begin{itemize}
\item \textbf{Exact recall lives in a few deep MLA layers.} For the \pairA{} seed (no token identity, needle 0.13 on
  this subset), making the single MLA layer 19 of Kimi-Linear native restores 100\% recall (+0.88
  \ci{+0.78}{+0.98}; layer 23: 0.80, layer 15: 0.58, others $\le$\,0.18), while making all 20 KDA states native does not
  help (0.03). The finalist's token channel fixes exactly these layers.
\item \textbf{Neither half suffices.} In both pairs, attention/MLA alone with empty states, and states alone with empty
  attention, lose every needle and raise val KL to 0.28--0.64. The recurrent state is not optional: with the
  finalist's maps but zero KDA state, \pairA{} CoQA F1 falls to 0.087 and ARC to 0.477; with zero GDN state the
  \pairB{} target loses the conversation entirely (237 of 300 CoQA answers start a new system turn).
\item \textbf{\pairB{}'s residual is half attention, half state.} All 16 attention layers native halve val KL
  (0.0159\,\fwd\,0.0078); all 48 GDN states native cut it by 38\% (\fwd\,0.0099). For the \pairA{} finalist no single
  layer or group explains the residual; the latents and the replayed states were trained together and only the full
  native set reaches the floor.
\end{itemize}

\subsection{Learning curves and \pairA{}'s overfitting to answer style}
\label{sec:curves}
\begin{figure}[t]
\centering
\includegraphics[width=\textwidth]{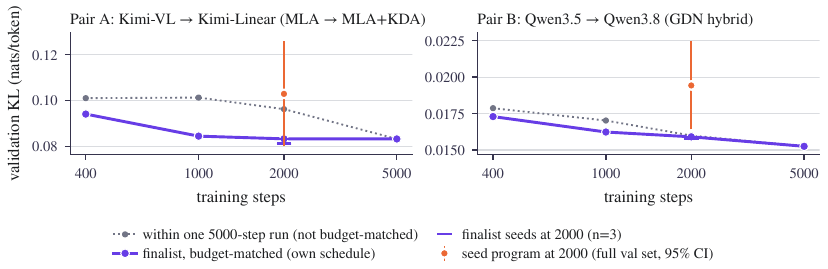}
\caption{Validation KL of the finalist against training budget. Solid: budget-matched (each point the end of its own
schedule); dotted: checkpoints inside one 5000-step run (warm-up / peak LR, not budget-matched); ticks at 2000: three
seeds.}
\label{fig:curves}
\end{figure}

Figure~\ref{fig:curves} shows validation KL against the training budget, each point the end of its own schedule
(budget-matched), plus checkpoints inside one 5000-step run.
\pairB{} improves monotonically but slowly in validation KL (0.0173 at 400 steps to 0.0153 at 5000) while task
metrics are at the native level from 400 steps on (CoQA F1 0.776\,\fwd\,0.788); 5000 steps (42
epochs over 950 conversations) show no overfitting on the held-out set.
\pairA{}'s val KL falls until about 2000 steps (0.094\,\fwd\,0.084\,\fwd\,0.082 at 400/1000/2000) and plateaus (0.083 at
5000), and needle recall is solved at every point. CoQA F1, however, falls from 0.558 (1000 steps, budget-matched) to
0.462 at 5000. Most of that fall is answer style: against the 1000-step checkpoint of the same run, F1 drops by 0.124
\ci{0.086}{0.160} with unchanged gold-in-answer (+0.000 \ci{$-$0.040}{+0.043}), and answers longer than 8 tokens rise
from 29\% to 51\% (target native: 11\%). Against the budget-matched 400- and 1000-step runs there is also a small real
content loss (gold-in-answer $-$0.037 \ci{$-$0.067}{$-$0.007} and $-$0.033 \ci{$-$0.067}{$-$0.003}). The translator is
trained only on chat continuations, so longer training moves the target towards long chat-style answers. On the full
sets, \pairA{} gains nothing from training beyond 1000--2000 steps.

\subsection{Cached objects are invertible: token identity and privacy}
\begin{figure}[t]
\centering
\includegraphics[width=\textwidth]{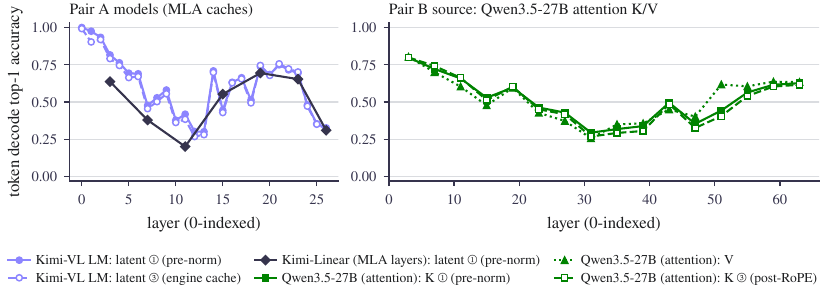}
\caption{Top-1 accuracy of decoding the token at each position from one layer's cached object (linear map + nearest
neighbour over the full vocabulary). Early MLA latents and attention keys identify the token.}
\label{fig:invert}
\end{figure}

Given one layer's cached per-token vector, a linear map to the token's input embedding followed by a nearest-neighbour
search over the full vocabulary recovers the token (Figure~\ref{fig:invert}; held-out tokens, a deliberately simple
decoder, so these are lower bounds). Kimi-VL's layer-0 MLA latent identifies the token with top-1 accuracy 1.000, also
for ids never seen in the fit, and layers 1--2 still 0.975 and 0.934. Every cached layer of every model tested
identifies most tokens with a linear read-out: at least 0.20 top-1 at the worst layer and typically 0.5--0.8, with a
dip in the middle of the stack. The engine-native \cpthree{} objects are as invertible as \cpone{} (within 0.03 per
layer). This is the lever behind the cache-only token-identity finalist, and it means a KV cache, MLA or GQA, is not an
anonymised representation of the prompt: a translated or stored cache should be protected like the prompt itself.

\subsection{Partial recompute (DroidSpeak-style baseline)}
\label{sec:partial}
\begin{figure}[t]
\centering
\includegraphics[width=\textwidth]{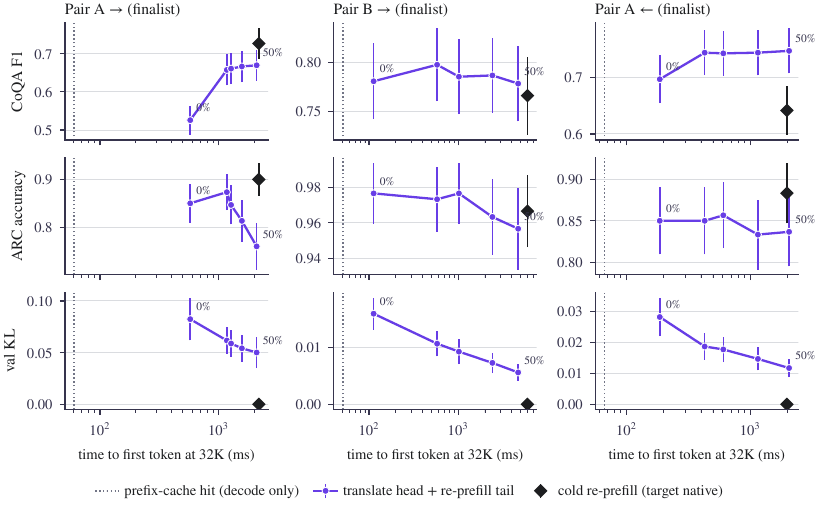}
\caption{Translating the head of the prompt and re-prefilling a tail fraction: quality vs time to first token at 32K,
against cold re-prefill (black) and a prefix-cache hit (dotted).}
\label{fig:partial}
\end{figure}

The target can receive the translated cache of the head of the prompt and re-prefill a tail fraction $f$ natively
(Figure~\ref{fig:partial}). A 5\% tail buys most of what the tail can buy: \pairA{} forward CoQA F1 0.526\,\fwd\,0.657
(+0.131 \ci{+0.098}{+0.166}; native 0.726), first-token KLs halve, \pairB{} val KL 0.0159\,\fwd\,0.0106. The cost
grows linearly with $f$, and for \pairA{} (3B active, cheap prefill) the 50\% point is as slow as re-prefill. For
\pairB{} the 5\% point still runs 10$\times$ faster than re-prefill at 32K (580 vs 6,000\,ms). ARC prompts are short, so
large tails split the question between translated and native states and cost accuracy (\pairA{} ARC $-$0.090 at 50\%).
Needle recall stays at 0.99--1.00 for every $f$.

\subsection{Model size and quality}
Across our pairs, the 27B dense pair (\pairB{}) transfers at the native level in both directions on every task metric
except the reverse refusals, the GQA 8B/4B pair at the small model's level, and the 3B-active MoE pair (\pairA{})
within the CI of native in the hybrid\,\fwd\,MLA direction but below it in the MLA\,\fwd\,hybrid direction. On domain
shift, \pairB{} keeps native
quality out of the chat distillation domain (GSM8K +0.035 \ci{0.000}{+0.075}; thinking-mode CoQA tie), while \pairA{}
forward drops 7.5 points on GSM8K to the source's own level (Appendix~\ref{app:breadth}). These pairs differ in
architecture, MoE vs dense and training-set size as well as in scale, so this is consistent with, not evidence for,
the hypothesis that larger receivers read translated caches better; a size-matched comparison would be needed.

%% file: frontier.tex
\section{Towards GLM, DeepSeek and Kimi flagships}
\label{sec:frontier}

\begin{figure}[t]
\centering
\includegraphics[width=\textwidth]{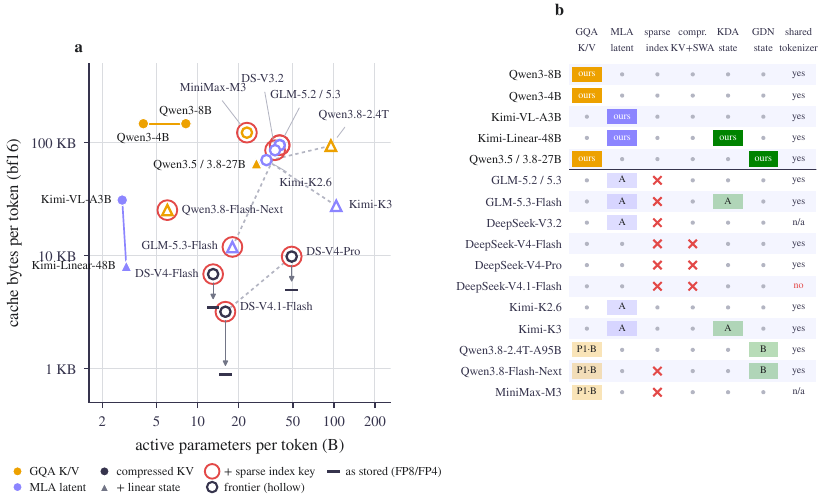}
\caption{Our pairs vs frontier models: active parameters vs cache bytes per token (left) and which cache objects each
model has (right). Configs pinned by commit on 29 Oct 2026.}
\label{fig:frontier}
\end{figure}

This section is an argument, not a result. It separates what we rehearsed from what is design only, and states the
scale-up as a hypothesis. Table~\ref{tab:evolution} (last column) and Figure~\ref{fig:frontier} place our pairs next to
the flagships, and \S\ref{sec:setup} lists where each pair differs from its flagship.

\paragraph{What transfers, as measured.}
Read against the flagship pairs of Table~\ref{tab:evolution}, our results say the following. \emph{Hybrid\,\fwd\,full
MLA} (GLM-5.3-Flash\,\fwd\,GLM-5.2/5.3; Kimi-K3\,\fwd\,Kimi-K2.6): from 7 latents plus token ids the target reached a
quality within the paired CI of its native level on every task metric (one seed), with exact recall to 64K and a
speed-up growing from $\times$2.7 at 1K to $\times$18.7 at 64K; the flagship hybrids hold the same inputs (11 and 24
latents). \emph{Full MLA\,\fwd\,hybrid}: exact recall needs token identity and states are synthesised by replay in
O($n$); this direction has the weakest results (judge $-$0.87, GSM8K $-$7.5 points, \cpthree{} fails beyond 8K, no gain
from training past 1--2K steps). \emph{Same-shape hybrid generations} (the Qwen3.8 family's GDN layers): copy plus a
closed-form correction gave native quality in both directions with an O(1) state transfer, apart from the inherited
refusals in the reverse direction. \emph{Multi-turn}: segment-wise transfer avoids the naive collapse on both pairs,
independently of the architecture.

\paragraph{Hypothesis (not established).}
These results support, but do not establish, that the per-object recipes are feasible for 300B+ hybrid flagships. The
per-token translation cost depends on the cached objects ($576\times576$ per MLA layer pair), not on the hidden size,
while prefill grows with active parameters, so the speed-up should grow with target size (\S\ref{sec:speed} states
the caveats, including the \pairA{} forward counter-case); a larger target may also read a translated cache better
(\pairB{} vs \pairA{} is consistent with this but confounded, \S\ref{sec:analysis}). Untested are the gaps of
\S\ref{sec:setup}: the scale gap (6--35$\times$ in active parameters), GLM's wider latent expansion (78/11 = 7.1
target latents per source latent vs our 27/7 = 3.9), the larger states to replay (3.4$\times$ ours for GLM-5.3-Flash,
10.3$\times$ for Kimi-K3), Kimi-K3's residual-stream changes, and DSA on MLA. By checkpoint size, three
sparse-attention models fit a 384\,GB node (DeepSeek-V4-Flash 160\,GB, Qwen3.8-Flash-Next-FP8 186\,GB, GLM-5.3-Flash
328\,GB); we rehearsed the indexer on the second.

\subsection{Rehearsal of sparse-attention indexer transfer}
\label{sec:dsa}
DSA-style models add a third per-token object: the indexer key that decides which tokens attention reads. A translator
can \emph{map} it (if the source has one), \emph{predict} it from what the target already receives, \emph{skip} it and
attend densely over the translated prefix (exact up to top-$k$, where sparse attention is dense anyway, but without the
decode savings), or combine dense and predicted (Appendix Figure~\ref{fig:dsadesign}). The risk is discrete: a small
key error can drop the needle's block out of the top-$k$ entirely.

We rehearsed this at small scale on one cross-model pair with one seed, \pairC{}: Qwen3.8-27B (GQA + GDN, no indexer)
\fwd{} Qwen3.8-Flash-Next (36 GDN + 12 QSA layers; QSA selects the top 512 blocks of 4 tokens, at most 2,051 tokens, by
a 128-d indexer key per token). The source has no indexer, so the keys are \emph{predicted} per layer from the
translated K and the target's token embedding, trained only through a KL on the target's native block-score
distribution (top-$k$ has no gradient); K/V maps go 16\,\fwd\,12 layers and GDN states 48\,\fwd\,36 layers by relative depth.
Our plumbing reproduces the native cache exactly (logit difference 0.0, identical selection and greedy tokens at 24 to
56,541 tokens).

\begin{figure}[t]
\centering
\includegraphics[width=\textwidth]{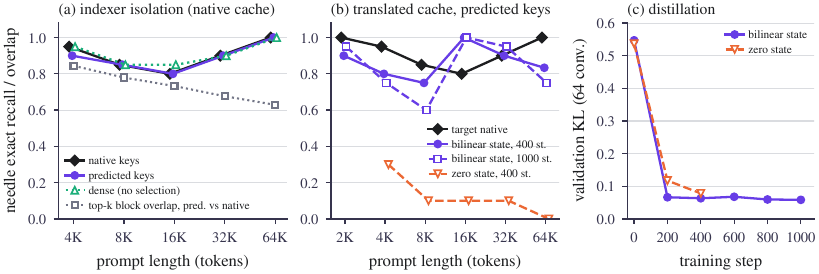}
\caption{\pairC{} (Qwen3.8-27B\,\fwd\,Qwen3.8-Flash-Next, QSA indexer), one seed. (a) On the target's own cache, predicted
indexer keys select only 63--84\% of the native top-$k$ blocks yet keep needle recall at the native level, as does dense
attention. (b) Translated cache decoded with predicted keys: with a bilinear-mapped GDN
state recall is near native at 400 and at 1000 steps; with zero state (400 steps) it is not. (c) Validation KL to
1000 steps (bilinear) and 400 (zero state); the bump at 600 follows the learning-rate restart at the resume from step 400.
$n{=}20$ per length at 2--32K, 12 at 64K (zero state: 4--64K only).}
\label{fig:dsa}
\end{figure}

\paragraph{The indexer is not the bottleneck on this pair.}
On the target's own cache with only the indexer keys replaced (Figure~\ref{fig:dsa}a), predicted keys (Stage-1
$R^2$ 0.60--0.72) select 84\% of the native blocks at 4K and 63\% at 64K, yet keep the needle block selected in at least
99.6\% of (layer, step) pairs and match native recall on 91 of 92 items; dense attention matches or exceeds native.
The discrete-selection failure did not appear because the needle block scores far above the budget boundary. This is
92 items on one model; it shows that the failure mode is not automatic, not that it cannot occur.

\paragraph{The state is the bottleneck.}
After 400 distillation steps (Figure~\ref{fig:dsa}b,c), the translator with a bilinear-mapped GDN state recalls 0.86 of
the 92 needles at 4--64K with predicted keys, against 0.87 with native keys, 0.80 dense, and 0.89 for the target
reading its own cache. With zero GDN state, validation KL is similar (0.078 vs 0.063) but recall is 0.13: validation
KL alone would have hidden the failure. At Stage 1 the closed-form K/V carried no needle at all (0 at every length):
unlike \pairB{}, the 27B and Flash-Next are different models, not same-architecture generations (K/V $R^2$ 0.38--0.73;
copying the 27B's states is harmful, val KL 3.90).

\paragraph{Longer distillation lowers KL but not recall.}
Continuing the bilinear run to 1000 steps (from step 400 on with the learning-rate schedule of a 1000-step run, hence
the bump at step 600) lowers validation KL from 0.063 to 0.058 (Figure~\ref{fig:dsa}c). Needle recall does not change
beyond one-seed noise: over 112 items at 2--64K, predicted keys recall 0.87 at 400 and 0.84 at 1000 steps, native keys
0.88 and 0.85, dense attention 0.82 and 0.85, against 0.91 for the target reading its own cache. The needle block stays
selected in at least 99\% of (layer, step) pairs at every length; native and predicted keys disagree on 3 of 112 items
at each checkpoint, in both directions, and in each of them the needle block was selected, so the different answer comes
from the other selected blocks. The per-length picture is uneven: at 16K the translated cache recalls 1.00 against the
target's own 0.80, but at 4K and 8K it stays below the target (predicted keys 0.80/0.75 at 400 steps and 0.75/0.60 at
1000, against 0.95/0.85 native), so the pooled tie hides a short-context deficit that one seed cannot resolve.
Absolute validation KL stays 3.6$\times$ above \pairB{}'s (0.058 vs 0.016 at 1000 steps, Figure~\ref{fig:curves}).

\paragraph{Task benchmarks.}
On \pairB{}'s CoQA, ARC and chat items (300/300/100), the 400- and 1000-step translators reach CoQA F1 0.795 and 0.765
(target 0.770), ARC 0.947 and 0.950 (target 0.980, three points above both; no paired CI was computed for this pair),
and chat token F1 against the target's own answers of 0.46 and 0.51 (chat first-token KL 1.06 and 0.89). A Gemma-4-31B
judge rates their chat answers 3.33 and 3.12 against 3.30 for the target's own answers; Flash-Next's own score of 3.30
is far below the 4.3--6.0 of the other targets, so the judge gap says little here. The closed form fails on everything
except ARC (CoQA 0.025, ARC 0.943, judge 2.53). With
\pairB{}'s score formula and this pair's own step-0 KL as reference, the combined score is 0.923 at 400 and 0.918 at
1000 steps (target 0.970, closed form 0.107), consistent with \pairB{}'s 0.887 and 0.894 at the same steps (checkpoints within one 5000-step run); with
\pairB{}'s KL reference it is 0.614 and 0.611, because the KL term then reflects the higher absolute KL of this
cross-model pair.

\vspace{3pt}
This is a rehearsal at small scale on one cross-model pair, one seed, up to 1000 steps; it does not test DSA on an MLA
cache (GLM-5.x, DeepSeek-V3.2), IndexShare, or FP8/MXFP4 indexer storage.

\paragraph{DeepSeek-V4/V4.1.}
No result of ours applies directly. A design would send token ids explicitly (the first layers are SWA-only or heavily
compressed, so ids cannot be read back from the cache), map compressed entries block-wise or recompute them through the
target's compressor, restore the 128-token SWA window with the publisher's own bounded recompute, and re-quantise.

%% file: limitations.tex
\section{Limitations}
\label{sec:limitations}

\begin{itemize}
\item \textbf{Scale.} Our pairs fit one machine (16--125B total, 3--27B active); the flagships are 6--35$\times$ larger
  in active parameters. Feasibility at 300B+ is a hypothesis (\S\ref{sec:frontier}); the speed extrapolation has no
  numeric FLOP model behind it.
\item \textbf{One tokenizer per family.} Every pair shares a tokenizer. Cross-family transfer (e.g.\ DeepSeek\,\fwd\,Kimi)
  needs a token aligner we have not validated.
\item \textbf{Sparse attention.} The indexer was rehearsed only on a QSA target (Qwen3.8-Flash-Next), one cross-model
  pair, one seed, up to 1000 steps, with 12--20 needle items per length; its combined score depends on the KL
  reference (0.92 with the pair's own step-0 KL, 0.61 with \pairB{}'s), the closed form's needle was not measured
  at 2K, and the translated cache stays below the target at 4K and 8K while the pooled recall ties it. The
  ``indexer is not the bottleneck'' finding rests on 92 isolation items and 112 translated-cache items on one model.
  DSA on MLA caches, IndexShare and low-precision indexer storage are not rehearsed.
  DeepSeek-V4's compressed multi-token entries are not rehearsed at all.
\item \textbf{\pairA{} forward is below native.} Chat judge $-$0.87 (5.15 vs 6.02), GSM8K $-$7.5 points, CoQA F1
  $-$0.20 (answer length), ARC $-$0.05; the engine-native capture \cpthree{} fails beyond the 8K training length; and
  training past 1--2K steps moves the answer style away from the target's.
\item \textbf{Behavioural leakage.} Translated states can carry the source's style and safety policy (\pairB{} reverse:
  6\% needle refusals); KL and accuracy do not detect it. The evidence is two markers on one pair.
\item \textbf{Seeds and coverage.} Part 1, the reverse translators' full-set rows, the multi-turn runs and \pairC{} are
  single-seed. The ``native level'' of \pairA{} reverse is a one-seed CI that includes 0 while gold-in-answer and ARC
  point 0.03--0.04 below native. The judge gap of \pairA{} forward on the deterministic re-run has no paired CI.
  \pairB{} reverse speed was not measured. The token-ID finalist of \pairA{} was not run in the multi-turn harness.
\item \textbf{Timing.} Speed is measured in Transformers, not in a serving engine; the \pairA{} cold arm carries an FP8
  dequantisation that we report separately. No flagship-size timing exists.
\item \textbf{Evaluation.} In-training subsets are the first items of the full probe sets; CoQA token F1 penalises
  answer length, so we report gold-in-answer next to it; the chat judge is a single LLM judge.
\end{itemize}

\section{Conclusion}
\label{sec:conclusion}
KV-cache translation extends from GQA to the cache objects of hybrid-attention models, but not with one mechanism.
MLA latents translate token-wise, best from the pre-norm, un-rotated capture; exact long-range recall after
synthesising a recurrent state needs token identity, which the MLA layer-0 latent supplies; same-geometry recurrent
states can be copied and corrected in closed form at O(1) cost; and sparse-attention indexer keys can be predicted
without losing the needle on the pair we tested. The hybrid\,\fwd\,full-MLA direction, which corresponds to handing a
conversation from a Flash hybrid to its full-MLA flagship, was not distinguishable from native quality on any task
metric, with speed-ups growing to $\times$18.7 at 64K. Whether these recipes hold at 300B+ scale is the hypothesis our evidence supports and the next
experiment should test.

%% file: appendix.tex
\begin{appendix}
\sloppy
\section{Build cost}
\label{app:cost}

\begin{table}[h]
\centering\small
\caption{Cost of building each translator on one GPU (from logs; \pairA{} forward capture time measured for
one shard and extrapolated by bytes). Amortisation point = build GPU-seconds / saved GPU-seconds per switch, from the
speed tables (not engine-adjusted; engine-adjusted \pairA{} forward: $\approx$74K / 18K switches).}
\label{tab:cost}
\begin{adjustbox}{max width=\linewidth}
\begin{tabular}{@{}lrrrrrr@{}}
\toprule
Direction & capture GPU-h & disk GB & Stage-2 GPU-h & params M & build GPU-h & amortised (8K / 32K) \\
\midrule
\pairA{} fwd cache-only & 0.16 & 262 & 7.7 & 102.0 & 8.0 & 42,731 / 15,626 \\
\pairA{} rev (a) & 0.61 & 104 & 1.5 & 91.1 & 2.1 & 24,417 / 3,369 \\
\pairB{} fwd bilinear & 0.20 & 228 & 3.5 & 110.1 & 3.7 & 8,792 / 2,036 \\
\pairB{} rev bilinear & 0.13 & 228 & 3.6 & 110.1 & 3.7 & -- \\
\bottomrule
\end{tabular}
\end{adjustbox}
\end{table}
\pairC{} trains at 82.6\,s per step on two GPUs (about 9.5\,h per 400-step run).

\section{Breadth: out-of-domain probes}
\label{app:breadth}
The distillation data are chat only, so these probes also test domain shift (Table~\ref{tab:breadth}; single seed,
finalists).

\begin{table}[h]
\centering\small
\caption{Out-of-domain probes, finalists, single seed: GSM8K accuracy (200 items, ``Answer: <number>'', 512 new tokens)
and thinking-mode CoQA (100 items, \pairB{} only; Kimi models have no thinking mode). Last column: finalist minus
target native, paired bootstrap 95\% CI.}
\label{tab:breadth}
\begin{adjustbox}{max width=\linewidth}
\begin{tabular}{@{}lcccc@{}}
\toprule
& target native & source native & finalist & finalist $-$ target \\
\midrule
\pairA{} GSM8K & 0.895 & 0.800 & 0.820 & $-$0.075 \ci{$-$0.125}{$-$0.025} \\
\pairB{} GSM8K & 0.830 & 0.871 ($n{=}196$) & 0.865 & +0.035 \ci{0.000}{+0.075} \\
\pairB{} CoQA thinking-on F1 & 0.753 & -- & 0.755 & +0.002 \ci{$-$0.024}{+0.026} \\
\pairB{} think block closed & 100\% & -- & 100\% & 0 \\
\bottomrule
\end{tabular}
\end{adjustbox}
\end{table}
\pairB{} transfers out of domain at the native level. \pairA{} forward lands at the source's own GSM8K level: on
multi-step arithmetic the translated cache gives the target roughly the source's competence, not its own.

\section{Long-context probes (\pairB)}
\label{app:longctx}
RULER-style~\citep{hsieh2024ruler} probes on Nemotron haystacks, 6 items per length (3 at 128K); no CI is meaningful
(Table~\ref{tab:longctx}).
\begin{table}[h]
\centering\small
\caption{\pairB{} forward long-context probes (single seed): multi-key needle and RULER-style variable tracking,
target native / bilinear finalist / seed (state copy), 6 items per length and 3 at 128K.}
\label{tab:longctx}
\begin{adjustbox}{max width=\linewidth}
\begin{tabular}{@{}lcccc@{}}
\toprule
& 8K & 32K & 64K & 128K \\
\midrule
multi-key needle exact (native / finalist / seed) & 1/1/1 & 1/1/1 & 1/1/1 & 1/1/1 \\
variable tracking, names found & .93/.97/.97 & 1.00/.97/.90 & .83/.97/.77 & .93/1.00/.87 \\
variable tracking, all exact & .67/.67/.33 & .67/.67/.50 & .33/.67/.17 & 0/0/0 \\
\bottomrule
\end{tabular}
\end{adjustbox}
\end{table}
Multi-key retrieval is saturated to 128K. Variable tracking is not saturated; the finalist is at or above the target at
every length and the plain state copy slightly lower at 32--128K. \pairA{} long-context needles are in
Table~\ref{tab:quality} (to 64K); \pairA{} was not evaluated at 128K.

\section{Further tables}
\label{app:tables}

\begin{table}[h]
\centering\small
\caption{\pairA{} reverse ladder at 400 steps, single seed (in-training subsets: CoQA 50, ARC 100, needle 8 per length;
Kimi-VL native 0.657 / 0.84 / 1.00). Variant (a) was carried to 2000 steps and is the reverse row of
Table~\ref{tab:quality}.}
\label{tab:ladder}
\begin{adjustbox}{max width=\linewidth}
\begin{tabular}{@{}lccccc@{}}
\toprule
Variant & params & val KL & CoQA F1 & ARC & needle \\
\midrule
closed form (a), step 0 & 91.1M & 0.113 & 0.578 & 0.74 & 0.43--0.67 \\
(a) token-wise & 91.1M & 0.0342 & 0.629 & 0.77 & 1.00 \\
(b) + causal local mixer & 104.1M & 0.0353 & 0.596 & 0.73 & 1.00 \\
(c) + mixer + KDA-state context & 115.3M & 0.0354 & 0.583 & 0.74 & 1.00 \\
(d) (a) + KDA-state context & 102.3M & 0.0343 & 0.604 & 0.74 & 1.00 \\
\bottomrule
\end{tabular}
\end{adjustbox}
\end{table}

\begin{table}[h]
\centering\small
\caption{Stage-1 $R^2$ and step-0 validation KL. \pairA{} reverse: held-out $R^2$ (mean over target layers 1--26) by
inputs; \pairB{}: forward vs reverse, and step-0 validation KL of the closed-form K/V maps under four state
treatments.}
\label{tab:r2}
\begin{adjustbox}{max width=\linewidth}
\begin{tabular}{@{}lc@{}}
\toprule
\pairA{} rev: token features only / nearest source latent + tok & 0.370 / 0.782 \\
\pairA{} rev: all 7 latents / all 7 + tok (variant a) & 0.821 / 0.834 \\
\pairB{} $R^2$ K / V, forward vs reverse & 0.772 / 0.716 vs 0.783 / 0.726 \\
\pairB{} step-0 val KL: zero / copy / bilinear / native states, fwd & 0.323 / 0.0595 / 0.0318 / 0.0129 \\
\pairB{} step-0 val KL, same, reverse & 0.315 / 0.0444 / 0.0286 / 0.0136 \\
\bottomrule
\end{tabular}
\end{adjustbox}
\end{table}

\begin{table}[h]
\centering\small
\caption{\pairA{} forward full sets: earlier run with the non-deterministic (atomic) MoE combine vs the deterministic
re-run used in the main text. All changes are within the earlier standard errors.}
\label{tab:olddet}
\begin{adjustbox}{max width=\linewidth}
\begin{tabular}{@{}lcccccc@{}}
\toprule
Anchor (old / det.) & val KL & CoQA F1 & CoQA gold & ARC & needle 2--32K & 64K \\
\midrule
target native & 0 / 0 & .721 / .726 & .803 / .803 & .903 / .900 & 1.00 / 1.00 & 1.00 / 1.00 \\
seed & .103 / .103 & .536 / .521 & .790 / .790 & .850 / .847 & 0.10 / 0.09 & 0.00 / 0.00 \\
cache-only & .083 / .082 & .519 / .526 & .817 / .813 & .850 / .850 & 0.99 / 0.99 & 0.83 / 0.83 \\
token-ID & .086 / .085 & .546 / .553 & .800 / .800 & .843 / .840 & 0.99 / 0.99 & 1.00 / 1.00 \\
\bottomrule
\end{tabular}
\end{adjustbox}
\end{table}

\begin{figure}[h]
\centering
\includegraphics[width=\textwidth]{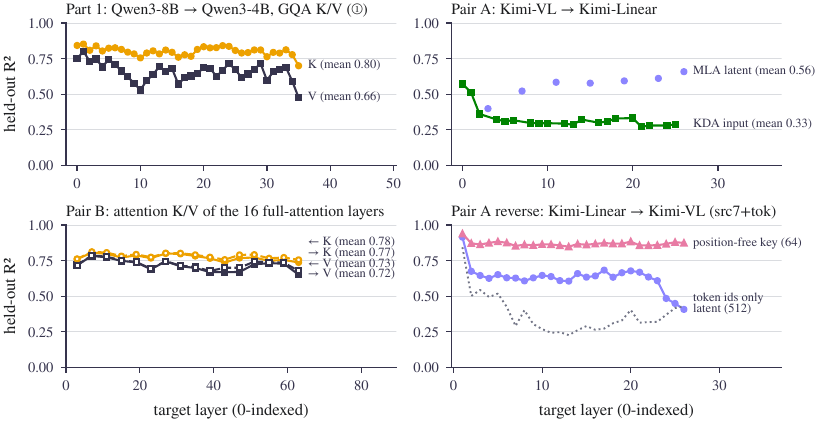}
\caption{Held-out $R^2$ of the closed-form Stage-1 maps per target layer (the share of the target object's variance the
map explains), all pairs and both directions.}
\label{fig:r2}
\end{figure}

\section{Effective rank of the maps}
\label{app:rank}
\begin{figure}[h]
\centering
\includegraphics[width=\textwidth]{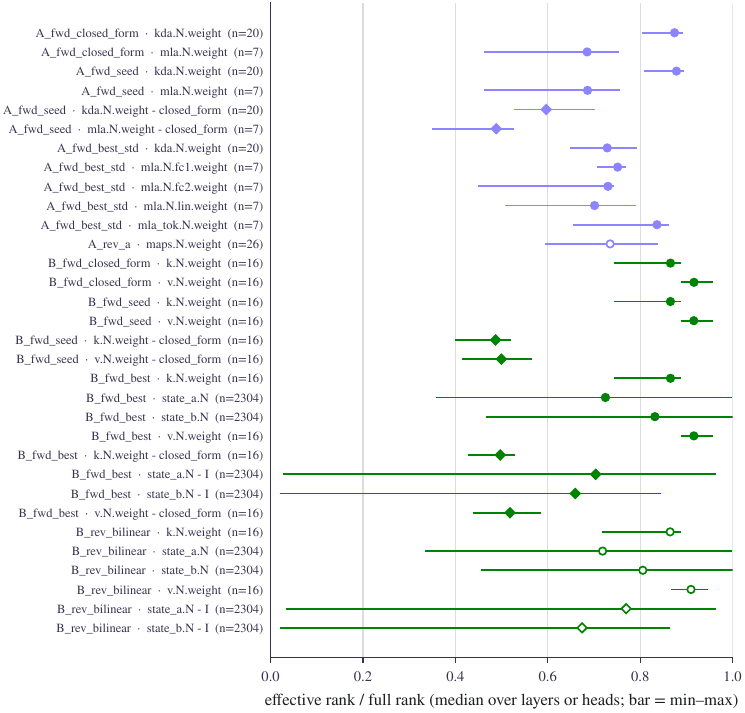}
\caption{Effective rank of every map family relative to its full rank, per translator. Diamonds: the trained change
relative to the closed-form initialisation (or identity for the bilinear state maps); hollow: reverse direction.}
\label{fig:rank}
\end{figure}

Per matrix we compute the entropy effective rank $\exp H(\sigma/\sum\sigma)$~\citep{roy2007erank}
(Figure~\ref{fig:rank}). The maps are high-rank, not low-rank adapters: the closed-form MLA maps of \pairA{} have a
median effective rank of 395 of 576, \pairA{} reverse 423 of 576, and \pairB{}'s K/V maps 886 / 938 of 1024. The Stage-2
update is small relative to the closed form (median $\|\Delta W\|/\|W_\mathrm{cf}\|$ 0.08 for the \pairA{} seed, 0.034
for the \pairB{} finalist) but spread over many directions. The bilinear GDN corrections are far from the identity in
most heads (median effective rank of $A-I$: 90 of 128).

\section{DSA transfer design}
\label{app:dsa}
\begin{figure}[h]
\centering
\includegraphics[width=\textwidth]{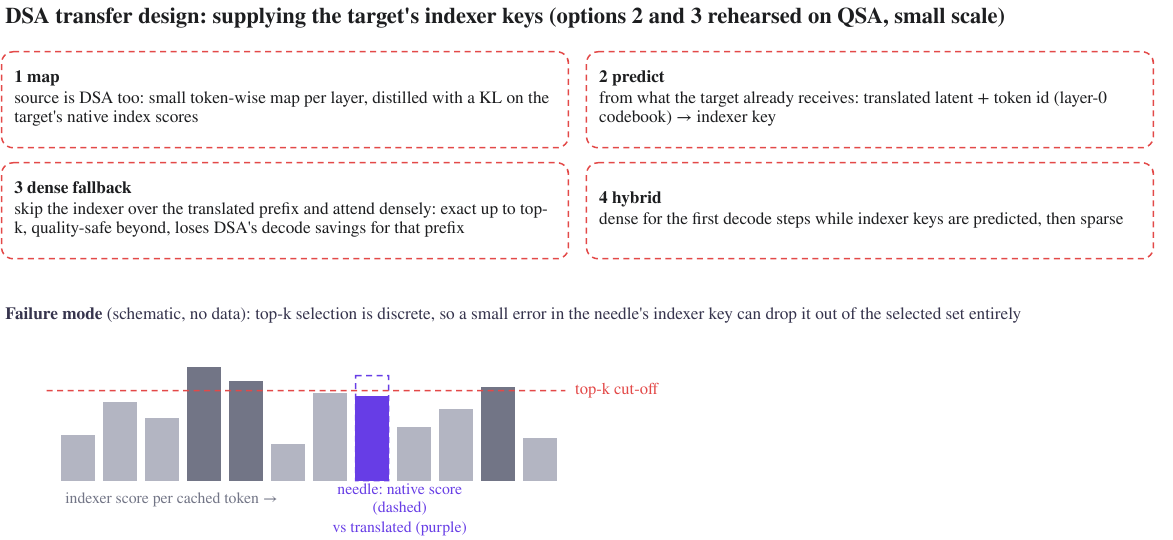}
\caption{Four ways to supply DSA indexer keys to a translated cache, and why discrete top-$k$ selection could make
needle recall fail all at once. Options 2 (predict) and 3 (dense) were rehearsed on a QSA pair (\S\ref{sec:dsa}).}
\label{fig:dsadesign}
\end{figure}

\FloatBarrier
\end{appendix}